%% file: main.tex
\documentclass[10pt,journal,compsoc]{IEEEtran}
\ifCLASSOPTIONcompsoc
  \usepackage[nocompress]{cite}
\else
  \usepackage{cite}
\fi

\ifCLASSINFOpdf
\else
\fi
\usepackage[utf8]{inputenc}

\usepackage{amsmath}       
\usepackage{amssymb}       
\usepackage{amsfonts}      
\usepackage{bm}            
\usepackage{amsthm}        
\usepackage{mathrsfs}      
\usepackage{microtype}     
\usepackage{xspace}        

\usepackage{graphicx}      
\usepackage{epsfig}        
\usepackage{epstopdf}      
\usepackage{xcolor}        
\usepackage{colortbl}      

\usepackage{array}         
\usepackage{booktabs}      
\usepackage{multirow}      
\usepackage{makecell}      
\usepackage{threeparttable}
\usepackage{adjustbox}     

\usepackage{enumitem}      

\usepackage{algorithm}     
\usepackage{algorithmic}   

\usepackage{url}           
\usepackage{balance}       
\usepackage{indentfirst}   
\usepackage{appendix}      
\usepackage{arydshln}      

\newcolumntype{I}{!{\vrule width 1.1pt}}

\begin{document}


\title{An Effective and Resilient Backdoor Attack Framework against Deep Neural Networks and Vision Transformers}

\author{
    
        Xueluan Gong*, Bowei Tian*,
        Meng Xue, Yuan Wu,
        Yanjiao Chen,~\IEEEmembership{Senior Member,~IEEE,}
        and
        Qian Wang,~\IEEEmembership{Fellow,~IEEE}

\IEEEcompsocitemizethanks{

\IEEEcompsocthanksitem X. Gong is with Nanyang Technological University, Singapore. E-mail: xueluan.gong@ntu.edu.sg

\IEEEcompsocthanksitem M. Xue is with the Department of Computer Science and Engineering, Hong Kong University of Science and Technology, China. E-mail: csexuemeng@ust.hk. 

\IEEEcompsocthanksitem Y. Wu is with the School of Computer Science and Artificial Intelligence \& Engineering Research Center of Hubei Province for Clothing Information, Wuhan Textile University. E-mail: wuyuanxu@whu.edu.cn.

\IEEEcompsocthanksitem Y. Chen is with the College of Electrical Engineering, Zhejiang University, China. E-mail: chenyanjiao@zju.edu.cn.

\IEEEcompsocthanksitem B. Tian and Q. Wang are with the School of Cyber Science and Engineering, Wuhan University, China. E-mail:boweitian@whu.edu.cn, qianwang@whu.edu.cn.

\IEEEcompsocthanksitem The first two authors have equal contributions.

%

}

}

\IEEEtitleabstractindextext{
\begin{abstract}

Recent studies have revealed the vulnerability of Deep Neural Network (DNN) models to backdoor attacks. However, existing backdoor attacks arbitrarily set the trigger mask or use a randomly selected trigger, which restricts the effectiveness and robustness of the generated backdoor triggers. In this paper, we propose a novel attention-based mask generation methodology that searches for the optimal trigger shape and location. We also introduce a Quality-of-Experience (QoE) term into the loss function and carefully adjust the transparency value of the trigger in order to make the backdoored samples to be more natural. To further improve the prediction accuracy of the victim model, we propose an alternating retraining algorithm in the backdoor injection process. The victim model is retrained with mixed poisoned datasets in even iterations and with only benign samples in odd iterations. Besides, we launch the backdoor attack under a co-optimized attack framework that alternately optimizes the backdoor trigger and backdoored model to further improve the attack performance. Apart from DNN models, we also extend our proposed attack method against vision transformers. We evaluate our proposed method with extensive experiments on VGG-Flower, CIFAR-10, GTSRB, CIFAR-100, and ImageNette datasets. It is shown that we can increase the attack success rate by as much as 82\% over baselines when the poison ratio is low and achieve a high QoE of the backdoored samples. Our proposed backdoor attack framework also showcases robustness against state-of-the-art backdoor defenses.

\end{abstract}


}
\maketitle

\IEEEdisplaynontitleabstractindextext

%
\IEEEpeerreviewmaketitle
\input{intro}

\ifCLASSOPTIONcaptionsoff
  \newpage
\fi

\end{document}

%% file: intro.tex
\IEEEraisesectionheading{\section{Introduction}}
\label{sec:intro}
\IEEEPARstart{D}{eep} neural networks have made tremendous progress in past years and are applied to a variety of real-world applications, such as face recognition \cite{schroff2015facenet}, automatic driving \cite{milz2018visual}, natural language processing \cite{mikolov2013distributed}, and objective detection \cite{redmon2016you}, due to superhuman performance.
Vision transformer (ViT) \cite{dosovitskiy2020image} is a promising deep learning architecture that offers a compelling alternative to traditional convolutional neural networks (CNNs) for computer vision applications. Despite the success in the computer vision domain, both DNN and ViT are vulnerable to backdoor attacks \cite{chen2020backdoor,gu2019badnets,lin2020composite,zheng2023trojvit,lv2021dbia}.
It is shown that the attacker can inject a backdoor (a.k.a. trojan) into the model by poisoning the training dataset during training time. The backdoored model behaves normally on the benign samples but predicts any sample attached with the backdoor trigger to a target false label. Due to its concealment, detecting backdoor attacks is very difficult. Moreover, the emergence of invisible backdoor triggers makes it more difficult to inspect whether the training samples are backdoored or not.

There exists a long line of backdoor attack strategies exploring injecting backdoors into DNNs \cite{gu2019badnets,liu2017trojaning, saha2019hidden,ji2017backdoor,ji2018model,lin2020composite,li2020rethinking,salem2020dynamic,yao2019latent,wang2020backdoor}.
However, they face the following shortcomings. First of all, most of the existing approaches \cite{gu2019badnets,liu2017trojaning} use a random backdoor trigger or random trigger mask (random pattern and location) in the attack, which is easy to be detected and achieves a sub-optimal attack performance. Second, current backdoor attacks \cite{gu2019badnets,liu2017trojaning, saha2019hidden,ji2017backdoor,ji2018model,li2020rethinking,salem2020dynamic,wang2020backdoor} separate the trigger generation process from the backdoor injection process, thus resulting in generating sub-optimal backdoor trigger and backdoored model. Third, various works utilize
visible backdoor triggers \cite{gu2019badnets, chen2017targeted, lin2020composite, nguyen2020input, liu2017trojaning, Gong2021, pang2020tale}, which can be easily detected by visual inspection.
Finally, although various existing works claimed to be defense-resistant \cite{Gong2021}, they can still be detected by the latest defenses, such as NAD \cite{li2021neuralv} and MNTD \cite{xu2019detecting}. In terms of backdoor attacks against ViTs, most of the existing transformer backdoor attacks use visible triggers to launch the attacks \cite{zheng2023trojvit,lv2021dbia}, making it easy for human defenders to detect abnormalities through visual inspections. Although Doan \cite{doan2023defending} proposed to generate hidden triggers based on a global warp of WaNet \cite{saha2019hidden}, the attack success rate and the perceptual trigger quality are relatively low.

In this paper, we put forward a novel backdoor attack strategy that integrates effectiveness and evasiveness. From the attack effectiveness perspective, unlike the existing works that use fixed trigger masks (e.g., a square in the lower right corner), we utilize an attention map to differentiate the weights of the pixels. The mask is determined as the pixels with the highest weights since such pixels have a higher impact on the classification. Using such a carefully designed trigger mask, we can achieve a higher attack success rate than the existing works with the same trigger size. Moreover, rather than separating the backdoor trigger generation from the backdoor injection process, we adopt the co-optimization backdoor framework that jointly optimizes the backdoor trigger and the backdoored model to generate an optimal backdoor trigger and achieve a higher attack success rate. In terms of evasiveness, it is quantified by both the human vision and state-of-the-art defense strategies. We carefully adjust the transparency (i.e., opacity) of the backdoor trigger and add a Quality-of-Experience (QoE) constraint to the loss function, aiming to generate a more natural backdoor trigger. Furthermore, we propose an alternating retraining algorithm that updates the model using either mixed samples or only clean samples according to the iteration index. In addition to evaluating DNN models, we also assess our proposed attack method on vision transformers. Experiments show that our proposed method outperforms baselines in both attack success rate and clean data accuracy, especially when the poison ratio is low. It is demonstrated that our proposed method is also robust to state-of-the-art backdoor defenses. 

{\color{black}This paper is an extended version of our previous paper \cite{Gong2022ATTEQNNAQ}, which is published in 2022 NDSS. 
We extend our previous work by extending the attack framework against vision transformers. While numerous studies have explored backdoor attacks against Convolutional Neural Networks (CNNs), there is a dearth of research on backdoor attacks tailored for vision transformers. Moreover, existing transformer backdoor attacks use visible triggers to launch the attacks \cite{zheng2023trojvit,lv2021dbia}, making it easy for human defenders to detect abnormalities through visual inspections. By extending our advanced backdoor attack framework to ViT models, we aim to drive the advancement of backdoor attacks. Due to the inherent differences between CNN and ViT architectures, it is not possible to directly transfer the CNN methodology to ViTs. In the Quality of Experience (QoE)-based trigger generation process, we analyze the ViT structure and select the head layer as the neuron-residing layer. To enhance the attack's effectiveness, we incorporate gradient enhancement techniques during trigger generation. We assign higher weights to the gradients of selected neurons that are critical for classifying the target label. This prioritization amplifies the poisoning effect of the generated trigger during the gradient descent optimization process. In addition, we conduct experiments to compare our proposed method with state-of-the-art ViT backdoor attacks, including DBIA \cite{lv2021dbia}, DBAVT \cite{doan2023defending}, BAVT \cite{subramanya2022backdoor}, and TrojViT \cite{zheng2023trojvit}. We also perform ablation studies to assess the effectiveness of different attack modules against ViT models. Furthermore, we demonstrate the resilience of our proposed attack against state-of-the-art ViT-specific backdoor defenses. }

To conclude, our paper makes the following contributions:
\begin{itemize}
    \item To the best of our knowledge, we are the first to utilize attention mechanisms to design backdoor trigger masks (i.e., trigger shape and trigger location), which significantly improves the attack performance. Rather than arbitrarily setting the mask, we determine the mask according to the focal area of the model to intensify the trigger impact on the prediction results.

    \item We propose a QoE-aware trigger generation method by introducing the QoE loss in the loss function to constrain the perceptual quality loss caused by the backdoor trigger.
    
    \item We design an alternating retraining method for backdoor injection to alleviate the decline of clean data prediction accuracy, which also helps resist state-of-the-art model-based defenses.
    
    \item Extensive experiments on VGG-Flower, GTSRB, CIFAR-10, CIFAR-100, and ImageNette datasets show that our proposed method outperforms the state-of-the-art backdoor attacks concerning both the attack effectiveness and evasiveness. We can evade state-of-the-art backdoor defenses. Apart from the DNN model, we show that our proposed attack method is also effective against vision transformers. 
    
\end{itemize}

\section{Background and Related work}

\subsection{Deep Neural Network}

Deep neural network is a class of machine learning models that uses nonlinear serial stacked processing layers to capture and model highly nonlinear data relationships. We mainly consider a prediction scenario, where a deep neural network $f_\theta$ encodes a function: $f_\theta: \mathcal{X} \rightarrow \mathcal{Y}$, $\theta$ is the parameter of $f$. Given the input sample $x  \in \mathcal{X}$, the DNN model $f_\theta$ outputs a nominal variable $f_\theta(x)$ ranging over a group of predesigned labels $\mathcal{Y}$.

The DNN model is usually trained by supervised learning. To obtain a DNN model $f$, the user utilizes a training dataset $\mathcal{D}$ that includes amounts of data pairs $(x,y) \in \mathcal{D} \subset \mathcal{X} \times \mathcal{Y}$, where $x$ is the input and $y$ is the ground-truth label of $x$. The trainer should determine the best $\theta$ for $f$ by optimizing the loss function $\mathcal{L}(f(x;\theta),y)$. The loss function is usually optimized by stochastic gradient descent \cite{amari1993backpropagation} and its derivatives \cite{zinkevich2010parallelized}.

However, training such sophisticated deep neural networks requires much computing and time costs since millions of parameters should be optimized. Therefore, many resource-limited clients prefer to outsource the training of deep neural networks to cloud computing providers, such as Google, Amazon, and Alibaba Cloud. Moreover, outsourcing training also has the following advantages. Firstly, optimizing the deep neural networks needs expert knowledge to determine the amenable model structure and much effort to fine-tune the hyperparameters. Second, training a sophisticated deep neural network requires millions of training samples. However, collecting and annotating them is labor-intensive for the clients. 
Based on the hindrance above, the cloud server provider receives more and more business of training DNN models. However, if the cloud providers are malicious, they may provide users with malicious models that will behave abnormally on specific samples. Being aware of such a threat, more and more defense works have been proposed to inspect whether the model is malicious. In this paper, we aim to design a more effective and defense-resistant backdoor attack methodology in the outsourced cloud environment from a malicious cloud server provider's perspective.

\subsection{Vision Transformer}

The Transformer architecture, initially designed for natural language processing (NLP) \cite{vaswani2017attention}, has been recently adapted for computer vision by leveraging the self-attention mechanism to model relationships between different parts of an image. One popular vision transformer is ViT \cite{dosovitskiy2020image}.

Let $X = \{x_1, x_2, ..., x_n\}$ be a sequence of $n$ input image patches, where each patch is represented as a tensor with dimensions $p \times p \times c$.
To begin, ViT applies an embedding layer to each image patch, transforming it into a $d$-dimensional embedding vector, which can be expressed as $E = \{e_1, e_2, ..., e_n\} = Embedding(X)$. Then, ViT employs a series of transformer encoder layers to process the embeddings. Each encoder layer consists of two sub-layers: a multi-head self-attention mechanism (MHSA) and a position-wise feedforward network (FFN).
The MHSA layer is responsible for capturing interactions between the patch embeddings using self-attention. The FFN layer applies a non-linear transformation to each patch embedding independently.

The attention mechanism within the Multi-Head Self-Attention (MHSA) layer can be divided into two main operations: attention rollout and attention diffusion. The attention rollout operation calculates the similarity between each query vector and all key vectors using the dot product. It scales the dot products by $\sqrt{d}$ to prevent the gradients from exploding, applies a softmax function to obtain attention weights, and finally computes a weighted sum of the value vectors. Mathematically, the attention rollout can be expressed as:
\begin{equation}
\text{Attention}(Q,K,V) = \text{softmax}\left(\frac{QK^T}{\sqrt{d}}\right)V,
\label{eq:1}
\end{equation}
where $Q$, $K$, and $V$ represent the query, key, and value matrices, respectively, and $d$ denotes the dimensionality of the key vectors.
The attention diffusion operation, on the other hand, can be expressed as follows:
\begin{equation}
\begin{split}
\text{MultiHead}(Q,K,V) = \text{Concat}\left(\text{head}_1,\ldots,\text{head}_h\right)W^O,\\
\text{head}_i = \text{Attention}(QW_i^Q,KW_i^K,VW_i^V),
\end{split}
\end{equation}
where $h$ represents the number of attention heads. $W_i^Q$, $W_i^K$, and $W_i^V$ are learnable weight matrices specific to the $i$-th attention head. $W^O$ is a learnable weight matrix used to map the concatenated output of all heads to the desired output dimensionality. The attention diffusion operation computes multiple attention heads in parallel and concatenates the resulting vectors along the last dimension. The concatenated vectors are then linearly transformed to obtain the final output.

In this paper, we also extend our proposed attack framework against vision transformers. Our experimental results demonstrate a high attack success rate when applied to vision transformers, highlighting the vulnerability of vision transformers to backdoor attacks.

\subsection{Backdoor Attacks against DNN Models}
In recent years, deep neural networks have been known to be vulnerable to backdoor attacks \cite{liu2017trojaning}. Intuitively, the objective of the backdoor attack is to trick the targeted DNN model into studying a powerful connection between the trigger and the target misclassification label by poisoning a small portion of the training dataset. As a result, every sample attached to the trigger will be misclassified to the target label with high confidence, while the backdoored model can also maintain high prediction accuracy on the benign inputs. 

To recap, the first backdoor attack is proposed by Gu et al. \cite{gu2019badnets}, namely BadNets. It is assumed that the attacker can control the training process of the DNN model. Thus, the attacker can poison the training dataset and change the configuration of the learning algorithms and even the model parameters. In BadNets, the attacker first chooses a random trigger (e.g., pixel perturbation) and poisons the training dataset with the backdoor trigger. After retraining the DNN model with the poisoned dataset, the DNN model will be backdoored. Based on the concept in BadNets, amounts of related works were proposed subsequently \cite{liu2017trojaning, saha2019hidden,ji2017backdoor,ji2018model,lin2020composite,li2020rethinking,salem2020dynamic,yao2019latent,wang2020backdoor}.

From the backdoor trigger perspective, rather than using the random trigger, Liu et al. proposed TrojanNN \cite{liu2017trojaning} that utilized a model-dependent trigger. The trigger is generated to maximize the activation of the selected neuron, in which the neuron has the largest sum of weights to the preceding layer. Further, considering to evade the pruning and retraining defenses, Wang et al. \cite{wang2020backdoor} put forward a ranking-based neuron selection methodology to choose neuron(s) that are difficult to be pruned and whose weights have little changes during the retraining process. Gong et al. \cite{Gong2021} selected the neuron that can be most activated by the samples of the targeted label to improve the attack strength.

Unlike using the above static backdoor triggers (i.e., fixed locations and patterns), Salem et al. \cite{salem2020dynamic} proposed a dynamic trigger generation strategy based on a generative network and demonstrated such dynamic triggers could evade the state-of-the-art defenses. Nguyen et al. \cite{nguyen2020input} implemented an input-aware trigger generator driven by diversity loss. A cross-trigger test is utilized to enforce trigger non-reusability, making it impossible to perform backdoor verification.

From the perspective of attack concealment, Saha et al. proposed hidden backdoor attacks \cite{saha2019hidden} in which the backdoored sample looks natural with the right labels. The key idea is to optimize the backdoored samples that are similar to the target images in the pixel space and similar to sourced images attached with the trigger in the feature space. Liao et al. \cite{liao2018backdoor} first generated an adversarial example that can alter the classification result and then used the pixel difference between the original sample and the adversarial example as the trigger. Li et al. \cite{li2019invisible} described the trigger generation as a bi-level optimization, where the backdoor trigger is optimized to enhance the activation of a group of neurons through L$_p$-regularization to achieve invisibility.

From the perspective of attack application scenarios, apart from targeting the centralized model, backdoor attacks against federated learning are also attracting much attention recently \cite{xie2019dba,bagdasaryan2018backdoor,liu2020backdoor, 10.1145/3243734.3243855,DBLP:conf/nips/WangSRVASLP20, gong2022coordinated,gong2022backdoor}. The attacker aims to backdoor the global model via manipulating his own local model. The main challenge is that the trigger will be diluted by subsequent benign updates quickly. In this paper, we only focus on backdoor attacks against centralized models.

Unlike the aforementioned backdoor attacks that either use ineffective random triggers or have visible triggers that can be easily detected, in this paper, we propose a more effective attention-based QoE-aware backdoor attack framework. It can not only achieve a high attack success rate but also evade state-of-the-art data-based backdoor defenses and human visual inspections.

\subsection{Backdoor Defenses for DNN Models}
When realizing the catastrophic impact of a backdoor attack, various defenses are also proposed to mitigate it. As far as we know, the exiting backdoor defense works can be categorized into data-based defense \cite{gao2019strip,chou2018sentinet,yang2020countermeasure,tran2018spectral,chen2018detecting,udeshi2019model} and model-based defense \cite{wang2019neural,liu2019abs,chen2018detecting,huang2019neuroninspect,chen2019deepinspect,li2021neural,gong2023redeem}. And both data-based and model-based defenses can also be further classified into online defense (during run-time) \cite{chen2020backdoor,gao2019strip,chou2018sentinet,yang2020countermeasure,liu2019abs,ma2019nic,udeshi2019model} and offline defense (before deployment) \cite{tran2018spectral,chen2018detecting,wang2019neural,huang2019neuroninspect,chen2019deepinspect}. 

Data-based backdoor defenses check whether a sample contains a trigger or not. From the perspective of online inspection, Gao et al. proposed Strip \cite{gao2019strip} that copies the inputting sample multiple times and combines each copy with a different sample to generate a novel perturbed sample. If the sample is benign, it is expected that those perturbed samples' prediction results will obtain a higher entropy result due to randomness. If the sample is backdoored, the prediction results will get a relatively low value since the trigger will strongly activate the targeted misclassification label. SentiNet \cite{chou2018sentinet} first seeks a contiguous region that is significant for the classification, and such region of the image is assumed to contain a trigger with high probability. Then SentiNet carves out the region, patches it on other images, and calculates the misclassification rate. If most of the patched samples are misclassified into the same false label, then the inputting sample is malicious. 
From the offline inspection perspective, Chen et al. proposed activation clustering, namely AC \cite{chen2018detecting}. It is known that the last hidden layer's activations can reflect high-level features used by the DNN to obtain the model prediction. AC assumes there exists a difference in target DNN activation between benign samples and backdoored samples with the same label. More concretely, if there exist backdoored samples in the inputs of a certain label, then the activation results will be clustered into two different clusters. And if the inputs contain no malicious samples, the activation cannot be clustered into distinct groups.
Tran et al. investigated spectral signature \cite{tran2018spectral}, which is based on statistical analysis, aiming to detect and eradicate 
malicious samples from a potentially poisoned dataset.

Model-based backdoor defenses check whether a deep neural network is backdoored or not. From the perspective of online inspection, Liu et al. \cite{liu2019abs} proposed Artificial Brain Stimulation (ABS) that is inspired by Electrical Brain Stimulation (EBS) to scan the target deep neural network and determine whether it is backdoored. 
Ma et al. proposed NIC \cite{ma2019nic} to detect malicious examples. NIC inspects both the provenance and activation value distribution channels.
From the offline inspection perspective, Wang et al. proposed Neural Cleanse (NC) \cite{wang2019neural} to inspect the DNN model. The key idea of NC is that as for the backdoored model, it needs much smaller modifications to make all input samples to be misclassified as the targeted false label than any other benign label.
Huang et al. proposed NeuronInspect \cite{huang2019neuroninspect} that integrates the output explanation with the outlier detection to reduce the detection cost. Chen et al. proposed DeepInspect \cite{chen2019deepinspect} that utilizes reverse engineering to reverse the training data. The key idea is to use a conditional generative model to get the probabilistic distribution of potential backdoor triggers. Xu et al. proposed MNTD \cite{xu2019detecting} that trains a meta-classifier to predict whether the model is backdoored or not. 

In this paper, we select a variety of representative defense works to defend our proposed attacks. It is shown that our proposed attack is robust to these defending works.

\subsection{Backdoor Attacks and Defenses against Vision Transformer}


{\color{black}To the best of our knowledge, the exploration of backdoor attacks against vision transformers is relatively limited, with only a few existing studies in this area. For instance, Lv et al. \cite{lv2021dbia} employed the attention mechanism of transformers to generate triggers and injected the backdoor by utilizing a poisoned surrogate dataset. Zheng et al. \cite{zheng2023trojvit} introduced TrojViT, which generates a patch-wise trigger to create a backdoor composed of vulnerable bits in the parameters of a vision transformer stored in DRAM memory. TrojViT achieves this through patch salience ranking and attention-target loss. Furthermore, TrojViT employs parameter distillation to minimize the number of vulnerable bits in the backdoor.

Recently, Yuan et al. \cite{yuan2023you} proposed BadViT, which leverages the self-attention mechanism in ViTs to manipulate the model's attention towards malicious patches. Additionally, the authors introduced an invisible variant of BadViT to increase the stealth of the attack by limiting the strength of the trigger perturbation. 
To improve backdoor stealth, several existing works have extended invisible CNN-oriented backdoor attacks to the ViT domain, such as BAVT \cite{subramanya2022backdoor} (built upon HB \cite{saha2019hidden}) and DBAVT \cite{doan2023defending} (built upon WaNet \cite{nguyen2021wanet}). However, these methods cannot consistently achieve a high attack success rate or maintain satisfying image quality.
}

To mitigate backdoor attacks on vision transformers, Subramanya et al. \cite{subramanya2022backdoor} presented a test-time defense strategy based on the interpretation map. Doan et al. \cite{doan2023defending} introduced a patch processing-based defense mechanism to mitigate backdoor attacks. The underlying idea behind these defenses is that the accuracy of clean data and the success rates of backdoor attacks on vision transformers exhibit different responses to patch transformations prior to the positional encoding. 

In this paper, we extend our proposed backdoor attack framework to vision transformers. It is shown that our proposed method outperforms the existing ViT-specific backdoor attacks regarding both effectiveness and evasiveness.

\section{Threat Model}
In this paper, we have the same threat model as the state-of-the-art backdoor attacks \cite{gu2019badnets,lin2020composite,salem2020dynamic}. We assume the attacker is a malicious cloud server provider responsible for training a sophisticated DNN/ViT for the clients. The attacker has the ability to control the model training process and access the training dataset. The training model structure, model parameters, and activation function are also transparent to the attackers. However, the attacker has no knowledge about the validation dataset that the clients use to test whether the received model is benign and satisfies the prediction accuracy. We also assume that the user is concerned about the security of the received model, i.e., he will inspect whether the model is backdoored using state-of-the-art defense strategies.

\section{Attack Methodology}

We first present the general attack framework and then describe key components in the framework, including attention-based mask determination, QoE-based trigger generation, and alternating retraining strategy. 

\subsection{Backdoor Attack Framework}
Since the attacker is capable of manipulating both the trigger and the model, we can formulate backdoor attacks as an optimization problem \cite{pang2020tale}.
\begin{equation}\label{equ:opt1}
    \min_{\delta, F_A} \mathcal{L}(x, F_V(x); F_A) + \lambda\mathcal{L} (x_t, y_t; F_A)  + \omega \mathcal{L}_{\delta}(x_t, x). 
\end{equation}
{\color{black}where $\mathcal{L}(\cdot)$ denotes the loss function and we have $\mathcal{L}_{\delta}(x_t, x)= ||x_t-x||_{\infty} = ||\delta||_{\infty}$. $\omega$ and $\lambda$ are constant parameters to balance the clean data accuracy and the attack success rate.}
The first term optimizes the prediction accuracy of clean samples. The second and third terms optimize the attack success rate of trigger-imposed samples while constraining trigger visibility. 

Optimizing (\ref{equ:opt1}) is difficult since the backdoor trigger $\delta$ and the backdoored model $F_A$ are co-dependent. Therefore, we separate the optimization problem (\ref{equ:opt1}) into two sub-problems and solve the two sub-problems by alternately updating the backdoor trigger $\delta$ and the backdoored model $F_A$ until convergence. We update the trigger and the model in the $k+1$-th iteration as 
\begin{equation}
\begin{split}
   \delta^{k+1} &= \arg\min_{\delta} \big(\mathcal{L}(x_t, F_A^{k}) + \omega \mathcal{L}_{\delta} (x_t, x)\big),\\
  F_A^{k+1} &= \arg\min_{F_A} \big( \mathcal{L}(x_t^{k+1}, F_A) + \lambda \mathcal{L} (x, F_V(x); F_A)\big).
  \end{split}
\end{equation}
Given the current model $F_A^k$, we first optimize the trigger $\delta^{k+1}$ using Adam optimizer \cite{kingma2014adam}, which will be elaborated in the following sections. Then, given the optimized trigger $\delta^{k+1}$, we obtain the optimized model $F_{A}^{k+1}$ by retraining the model $F_A^k$ with poisoned samples using $\delta^{k+1}$. 
We summarize the algorithm of the co-optimization attack framework in Algorithm~\ref{alg:backdoor}.

\begin{algorithm}[tt]
	\caption{Attention-based QoE-aware backdoor attack.} \label{alg:backdoor}
	\begin{algorithmic}[1]
		\REQUIRE Pre-trained benign deep neural network $F_V$, trigger size $l^2$, target label $y_t$, training samples $\mathcal{D}$, parameters $\lambda, \omega$.
		\ENSURE Trigger $\delta$, backdoored model $F_A$.
		\STATE // Attention-based mask generation
		\STATE $H_{opt}(x) = RAN(\mathcal{X}_t)$.
		\STATE Select $l^2$ pixels with the highest weight in $H_{opt}(x)$ to form $M$.
		\STATE // Initialize the trigger and the model
		\STATE $k=0$.
		\STATE $\delta^k = Mask\_Initialize(M)$.
		\STATE $F_A^k$ = $F_V$.
		\WHILE {\emph{not convergence}}
		\STATE $k = k + 1 $.
		\STATE // QoE-aware trigger generation
		\STATE $\delta^{k} = Trigger\_Optimize(F_A^{k-1}, \lambda, \mathcal{D}, SSIM)$.
		\STATE // Alternating retraining for backdoor injection
		\STATE The retraining dataset $\mathcal{D}_r = Alt\_Retrain(k, \mathcal{D}, \delta^{k})$. 
	    \STATE $F_A^{k} = Model\_Retrain(F_A^{k-1}, \delta^{k}, \omega,  \mathcal{D}_r)$.
	    \ENDWHILE
	    \RETURN $\delta^k$ and $F_A^k$.
  \end{algorithmic}
\end{algorithm}

\subsection{Attention-based Mask Determination}
{\color{black} In classification tasks, the classification model focuses on different parts of the input image, similar to the human visual system. For a specific class (e.g., deer), most high-performing classification models of different architectures usually pay attention to the same key features (e.g., antlers), as demonstrated by numerous research works on explaining machine learning models using attention networks \cite{vaswani2017attention,huang2020attributes,dosovitskiy2020image}. Manipulating the pixels of high importance is more likely to divert the classification results.

Unlike CNNs, which rely on spatial hierarchies to extract features, ViTs break down images into patches and use self-attention to weigh the contribution of each patch in the classification
process. By pinpointing and strategically altering the patches that have the most significant influence on the model’s output, attackers can hijack the model’s decision-making, leading to a higher likelihood of misclassification. Besides, since ViTs lack inherent hierarchical feature abstraction, they are more susceptible to input perturbations amplified by the attention mechanism. Thus, altering attention weights in key feature patches can misdirect the model’s focus, resulting in misclassification. 

Motivated by this, we propose an attention-based trigger mask determination method to select the most significant pixels as the trigger mask. This approach generates powerful triggers that achieve better attack performance.} 
{\color{black}In this paper, we utilize a residual attention network (RAN) \cite{wang2017residual} to obtain attention maps for both DNN and ViT models.} {\color{black}RAN is a feed-forward CNN with stacks of attention modules to extract the features for classification in the residual network. Each attention module consists of a trunk branch $T$ and a soft mask branch $S$. The trunk branch processes features of neural networks, and the soft mask branch selects features by imitating the human cortex path \cite{mnih2014recurrent}. RAN combines bottom-up and top-down learning methods to realize fast feed-forward processing and top-down attention feedback in one feed-forward procedure.}

{\color{black}An input sample $x_i$ first passes through a residual unit to get $x_i^1$ as the input to the first attention module.} In a RAN with $L$ attention modules, the output of the $l$-th attention module is
\begin{equation}
    H_{l,c} (x_i^{l}) = (1+S_{l,c}(x_i^l)) \cdot T_{l,c}(x_i^l), c \in [1,2,...,C_l],
\end{equation}
{\color{black}where $S_{l,c}(\cdot)$ and $T_{l,c}(\cdot)$ are the $c$-th channel of the mask branch and the trunk branch of the $l$-th attention module respectively, and $C_l$ is the number of channels in the $l$-th attention module. The output $H_{l,c}$ will be fed into the $l+1$-th attention module after a residual unit. 

In RAN, different attention modules play different roles. Low-level attention modules reduce the influence of unimportant background features, and high-level attention modules pick up important features that enhance classification performance. The output of the final attention module is the attention map with attention weights for corresponding pixels. The attention weights represent the degree of attention the model pays to each pixel, reflecting the contribution of each pixel to driving the prediction results of the image into a certain class.

The size of the obtained attention map is the same as the size of the output of RAN, which may be different from the size of the input. For instance, in our experiments, given a $32 \times 32$ image, the size of the output of the last attention module is $8\times 8$, which is smaller than the input size. We upscale the attention maps to the same size as the input by bilinear interpolation \cite{kirkland2010bilinear}. We use $H(x_i)$ to denote the upscaled attention map of sample $x_i$.}

We randomly select $N$ clean samples of the target class $y_t$ and attain $N$ attention maps $\{{H(x_i)}\}_{i=1}^N$. Assuming that each sample has the same probability of occurrence, we choose the attention map that is closest to the average attention map for generality. 
\begin{equation}
\begin{split}
    H_{opt}(x)= \arg\min_{x_i \in \mathcal{X}_t} || \bar{H}(x)-H(x_i)||_2,
\end{split}
\end{equation}
where $\mathcal{X}_t$ is the set of samples of the target label $y_t$, and $\bar{H}(x) = \frac{\sum_{j=1}^{j=N} H(x_j)}{N}$ is the average attention map. 

Considering that most existing works use a contiguous square trigger of size $l \times l$ ($l$ is the number of pixels), we also use the conventional expression $l \times l$ to denote the trigger size. To make a fair comparison, we choose the top $l^2$ pixels with the highest attention values as the trigger region, i.e., trigger mask $M$, in our attack for evaluation. 

\begin{table*}[tt]
	\caption{Comparison of our proposed attack framework with state-of-the-art DNN-specific backdoor attacks for VGG-Flower-l, CIFAR-10, and GTSRB. }
	\label{tab:com1}
	\centering
	\footnotesize
	\begin{tabular}{c|cccccccccc}
		\toprule
		&\multicolumn{10}{c}{VGG-Flower-l}   \\
		\multirow{2}{*}{\shortstack{\#ratio}}
		&\multicolumn{2}{c}{BadNets \cite{gu2019badnets}} & \multicolumn{2}{c}{TrojanNN \cite{liu2017trojaning}} &\multicolumn{2}{c}{HB \cite{saha2019hidden}}&\multicolumn{2}{c}{RobNet \cite{Gong2021}}
		&\multicolumn{2}{c}{Ours}\\ & ASR &CDA  & ASR &CDA  & ASR &CDA& ASR &CDA& ASR &CDA \\
		\midrule
		$10\%$ &22.00\%&96.0\%&21.00\%&94.50\%&19.00\%&94.0\%&82.50\%& 95.50\%&\textbf{94.50\%} &\textbf{97.00\%}   \\
		$15\%$ &22.50\%&95.00\%&22.00\%&95.50\%&24.00\%&93.50\%&80.50\%&92.50\% &\textbf{99.00\%} & \textbf{96.00\%}   \\
		$20\%$ &22.50\%&96.50\%&23.00\%&96.50\%&22.00\%&94.50\%&89.50\%& 91.50\% & \textbf{99.00\%} & \textbf{97.50\%}  \\
		$25\%$ &24.50\%&94.50\%&27.00\%&93.00\%&33.00\%&95.00\%&91.00\% & 96.00\% &\textbf{100.0\%} & \textbf{98.00\%}   \\
		$30\%$ &26.50\%&97.00\%&27.50\%&94.00\%&36.50\%&95.00\%&99.50\% & 95.00\% &\textbf{100.0\%} & \textbf{98.50\%}  \\
		\hline
		&\multicolumn{10}{c}{CIFAR-10}   \\
		\multirow{2}{*}{\shortstack{\#ratio}}
		&\multicolumn{2}{c}{BadNets \cite{gu2019badnets}} & \multicolumn{2}{c}{TrojanNN \cite{liu2017trojaning}} &\multicolumn{2}{c}{HB \cite{saha2019hidden}}&\multicolumn{2}{c}{RobNet \cite{Gong2021}}&\multicolumn{2}{c}{Ours}\\ & ASR &CDA & ASR &CDA & ASR &CDA  & ASR &CDA& ASR &CDA \\
		\midrule
		$1\%$ &10.00\%&87.98\% & 11.82\% &85.95\% & 27.83\% & 88.02\% &32.7\% & 87.92\% & \textbf{44.69\%} & \textbf{88.98\%}\\
		$3\%$ &10.34\% & 90.92\% & 12.28\% & \textbf{90.99\%} & 31.24\% &87.55\% & 65.79\% & 88.23\%&\textbf{86.84\%} & 88.35\%\\
		$5\%$ & 93.93\% & 90.02\% & 97.09\% & 89.87\% & 30.07\% & \textbf{90.03\%}& 95.62\% & 88.39\% & \textbf{97.29\%} & 88.90\% \\ 
		$10\%$ & 95.43\% & 88.90\% & 98.05\% &89.67\% & 29.07\% & 85.22\% & 95.06\% & 87.84\% & \textbf{99.26\%} & \textbf{90.10\%}\\ 
		$15\%$ & 97.06\% & 88.32\% & 98.77\% & 87.69\% & 44.74\% & 84.89\% & 96.30\% & 87.65\% & \textbf{99.33\%} & \textbf{89.12\%}\\ 
		$20\%$ & 98.06\% & 89.54\% & \textbf{99.75\%} & 85.20\% & 60.08\% &86.07\% & 96.93\% & 87.64\%&99.01\% & \textbf{90.07\%} \\
		\hline
		&\multicolumn{10}{c}{GTSRB}   \\
		\multirow{2}{*}{\shortstack{\#ratio}}
		&\multicolumn{2}{c}{BadNets \cite{gu2019badnets}} & \multicolumn{2}{c}{TrojanNN \cite{liu2017trojaning}} &\multicolumn{2}{c}{HB \cite{saha2019hidden}}&\multicolumn{2}{c}{RobNet \cite{Gong2021}}&\multicolumn{2}{c}{Ours}\\ & ASR &CDA & ASR &CDA & ASR &CDA  & ASR &CDA& ASR &CDA \\
		\midrule
		$0.3\%$ &22.01\%&92.57\%&25.55\%&94.14\%&8.01\%&89.43\%&26.81\%& 96.60\%
		&\textbf{90.88\%}&\textbf{97.15\%}\\
		$0.5\%$ &46.52\%&94.09\%&47.50\%&95.38\%&12.01\%&90.08\%&56.99\%& 96.89\%
		&\textbf{93.30\%}&\textbf{97.08\%}\\ 
		$1\%$ &96.25\% &94.46\%&96.65\%&94.98\%&23.60\%&89.07\%& \textbf{98.84\% }& 95.53\%
		& 96.75\% & \textbf{96.94\%}\\
		$3\%$ &97.81\% &95.00\%&97.93\%&94.46\%&77.25\%&89.67\%& \textbf{99.95\%}& 96.80\%
		&99.39\% & \textbf{97.11\%}\\
		$5\%$ &98.08\% &96.22\%&98.10\%&96.25\%&77.74\%&88.10\%& 99.51\% & \textbf{97.36\%} 
		&\textbf{99.97\%} & 97.19\%\\ 
		$7\%$ &98.91\% &96.54\%&98.97\%&95.76\%&78.27\%&88.31\%& 99.20\% & 96.04\% 
		&\textbf{99.91\%}& \textbf{96.81\%}\\ 
	    
		\bottomrule
	\end{tabular}
	
\end{table*}

\begin{figure*}[tt]
	\centering
	\hspace{-0cm}
	\begin{minipage}[t]{3.3in}
		\centering
		\includegraphics[trim=0mm 0mm 0mm 0mm, clip,width=3.3in]{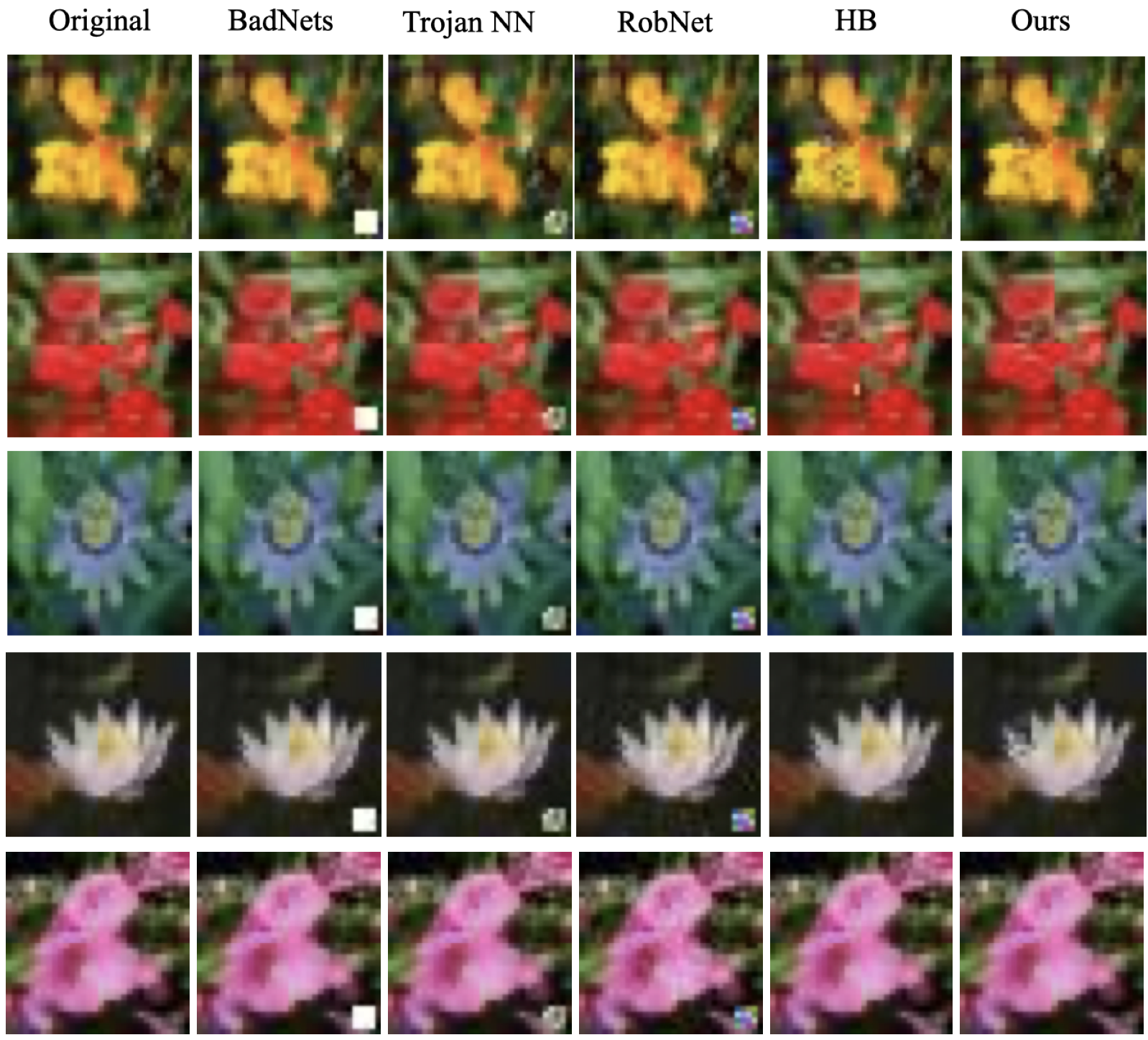}
		\centerline{\footnotesize VGG-Flower-l}
	\end{minipage}
	\begin{minipage}[t]{3.3in}
		\centering
		\includegraphics[trim=0mm 0mm 0mm 0mm, clip,width=3.3in]{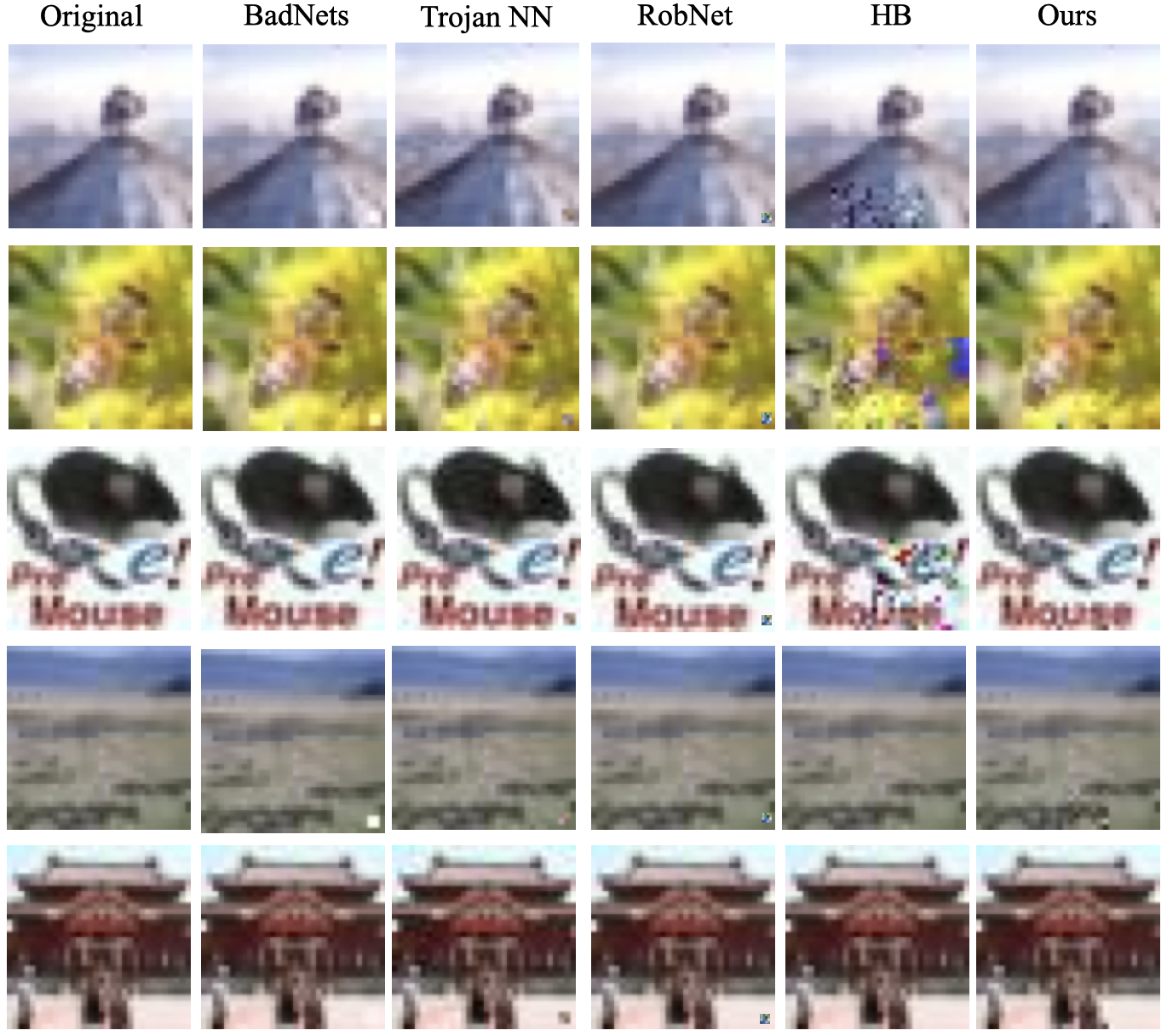}\\
		\centerline{\footnotesize CIFAR-100}
	\end{minipage}
	\caption{Comparison of backdoored samples between our method and the baselines against CNNs.} \label{fig:poisoned}
\end{figure*}

\begin{table*}[tt]
	\caption{Comparison of our proposed attack framework with state-of-the-art DNN-specific backdoor attacks for CIFAR-100, ImageNette, and VGG-Flower-h. }
	\label{tab:com2}
	\centering
	\footnotesize
	\begin{tabular}{c|cccccccccc}
		\toprule
		
		&\multicolumn{10}{c}{CIFAR-100}   \\
		\multirow{2}{*}{\shortstack{\#ratio}}
		&\multicolumn{2}{c}{BadNets \cite{gu2019badnets}} & \multicolumn{2}{c}{TrojanNN \cite{liu2017trojaning}} &\multicolumn{2}{c}{HB \cite{saha2019hidden}}&\multicolumn{2}{c}{RobNet \cite{Gong2021}}&\multicolumn{2}{c}{Ours}\\ & ASR &CDA& ASR &CDA  & ASR &CDA  & ASR &CDA& ASR &CDA \\
		\midrule
		$0.1\%$ & 1.29\% & 73.57\% & 1.61\% & 74.66\% &3.04\% &68.79\% &17.01\% & 73.02\% &\textbf{96.53\%} & \textbf{74.55\%}\\
		$0.3\%$ &2.52\% & 73.48\% & 2.25\% & 74.28\% &3.88\% &69.52\% & 98.01\%& 71.45\% & \textbf{98.66\%} & \textbf{75.06\%}\\
		$0.5\%$ &2.47\% & 73.08\% & 2.5\% & 73.62\% &3.68\% & 67.03\%& 97.33\% & 71.67\% &\textbf{99.94\%} & \textbf{74.91\%}\\ 
		$1\%$   & 2.56\% & 73.36\%&3.27\% & 72.99\% &7.44\%&69.94\% &98.66\% & 71.72\% & \textbf{99.78\%} & \textbf{74.64\%} \\ 
		$3\%$   &90.38\%& 71.59\% & 95.61\% & 73.13\% &62.73\% &70.28\% &99.49\% & 72.44\% & \textbf{99.84\%} & \textbf{75.44\%}\\
		 \hline
		&\multicolumn{10}{c}{\color{black}ImageNette}   \\
		\multirow{2}{*}{\shortstack{\color{black}\# ratio}}
		&\multicolumn{2}{c}{\color{black}BadNets \cite{gu2019badnets}} & \multicolumn{2}{c}{\color{black}TrojanNN \cite{liu2017trojaning}} &\multicolumn{2}{c}{\color{black}HB \cite{saha2019hidden}}&\multicolumn{2}{c}{\color{black}RobNet \cite{Gong2021}}&\multicolumn{2}{c}{\color{black}Ours}\\ & \color{black}ASR &\color{black}CDA  & \color{black}ASR &\color{black}CDA  & \color{black}ASR &\color{black}CDA& \color{black}ASR &\color{black}CDA& \color{black}ASR &\color{black}CDA \\
		\midrule
		{\color{black}$5\%$} &\color{black}11.52\%&\color{black}91.49\%&\color{black}11.41\%&\color{black}91.77\%&\color{black}10.60\%&\color{black}91.40\%&\color{black}60.31\%&\color{black}88.96\%&\color{black}\textbf{88.82\%}&\color{black}\textbf{91.95\%} \\
		\color{black}$10\%$&\color{black}13.45\%&\color{black}90.50\%&\color{black}14.15\%&\color{black}90.24\%&\color{black}11.81\%&\color{black}89.93\%&\color{black}68.78\% &\color{black}86.42\%&\color{black}\textbf{90.83\%}&\color{black}\textbf{90.59\%}\\
		\color{black}$15\%$&\color{black}14.26\%&\color{black}89.00\%&\color{black}15.28\%&\color{black}88.14\%&\color{black}14.42\%&\color{black}91.40\%&\color{black}81.98\%&\color{black}88.82\%&\color{black}\textbf{92.16\%}&\color{black}\textbf{92.40\%}\\ 
		\color{black}$20\%$   &\color{black}21.53\%&\color{black}86.50\%&\color{black}24.89\%&\color{black}85.83\%&\color{black}15.34\%&\color{black}88.27\%&\color{black}85.50\%&\color{black}88.16\%&\color{black}\textbf{95.01\%
		}&\color{black}\textbf{91.57\%}\\ 
	\color{black}	$30\%$&\color{black}35.13\%&\color{black}71.28\%&\color{black}37.83\%&\color{black}70.54\%&\color{black}18.81\%&\color{black}85.32\%&\color{black}92.92\%&\color{black}84.87\%&\color{black}\textbf{97.58\%}&\color{black}\textbf{91.46\%}\\
		 \hline
		&\multicolumn{10}{c}{\color{black}VGG-Flower-h}   \\
		\multirow{2}{*}{\shortstack{\color{black}\#ratio}}
		&\multicolumn{2}{c}{\color{black}BadNets \cite{gu2019badnets}} & \multicolumn{2}{c}{\color{black}TrojanNN \cite{liu2017trojaning}} &\multicolumn{2}{c}{\color{black}HB \cite{saha2019hidden}}&\multicolumn{2}{c}{\color{black}RobNet \cite{Gong2021}}&\multicolumn{2}{c}{\color{black}Ours}\\ & \color{black}ASR &\color{black}CDA  & \color{black}ASR &\color{black}CDA  & \color{black}ASR &\color{black}CDA& \color{black}ASR &\color{black}CDA& \color{black}ASR &\color{black}CDA \\
		\midrule
	\color{black}	$10\%$ &\color{black}11.00\%&\color{black}95.50\%&\color{black}12.50\%&\color{black}95.50\%&\color{black}5.50\%&\color{black}96.00\%&{\color{black}34.50\%}&\color{black}95.00\%&\color{black}\textbf{40.00\%}&\color{black}\textbf{98.50\%} \\
		\color{black}$15\%$ &\color{black}15.50\%&\color{black}95.50\%&\color{black}16.50\%&\color{black}95.00\%&\color{black}6.50\%&\color{black}95.00\%&\color{black}58.50\%&\color{black}95.00\%&\color{black}\textbf{83.00\%}&\color{black}\textbf{96.00\%}\\
	\color{black}	$20\%$ &\color{black}20.00\%&\color{black}94.00\%&\color{black}23.00\%&\color{black}96.50\%&\color{black}15.50\%&\color{black}95.50\%&\color{black}60.00\%&\color{black}97.00\%&\color{black}\textbf{92.50\%}&\color{black}\textbf{97.50\%}\\ 
	\color{black}	$25\%$ &\color{black}28.00\%&\color{black}96.50\%&\color{black}27.00\%&\color{black}94.50\%&\color{black}19.50\%&\color{black}94.50\%&\color{black}73.00\%&\color{black}94.00\%&\color{black}\textbf{98.50\%}&\color{black}\textbf{97.50\%}\\ 
	\color{black}	$30\%$   &\color{black}30.00\%&\color{black}95.50\%&\color{black}29.00\%&\color{black}95.50\%&\color{black}21.50\%&\color{black}93.00\%&\color{black}76.50\%&\color{black}95.50\%&\color{black}\textbf{100.0\%}&\color{black}\textbf{97.00\%}\\
		\bottomrule
	\end{tabular}
	
\end{table*}

\begin{table*}[tt]
	\caption{{\color{black}Compare our proposed method with state-of-the-art ViT-specific backdoor attacks.}}
	\label{tab:Comvit}
	\centering
	\setlength\tabcolsep{12pt}
	\footnotesize
	\begin{tabular}{cc|cccccc}
		\toprule
		Datasets&Metrics&DBIA \cite{lv2021dbia} &DBAVT \cite{doan2023defending}& BAVT \cite{subramanya2022backdoor}&{\color{black}TrojViT \cite{zheng2023trojvit}}&Ours \\
		\midrule
		\multirow{4}{*}{\shortstack{VGG-Flower-l}}&ASR&90.68\%&95.02\%&77.10\%&\color{black}94.90\%&\textbf{95.70\%}\\
             &CDA&95.98\%&94.05\%&80.30\%&\color{black}97.12\%&\textbf{99.00\%}\\
             &SSIM&0.9504&0.9910&0.9995&\color{black}0.9102
             &\textbf{0.9989}\\
             &LPIPS&0.1523&0.0403&0.0867&\color{black}0.5568&\textbf{0.0122}&\\
             \hline
	    \multirow{4}{*}{\shortstack{CIFAR-10}}&ASR&98.40\%&96.00\%&80.30\%&\color{black}98.74\%&\textbf{98.89\%}\\
             &CDA&96.32\%&98.00\%&84.50\%&\color{black}98.66\%&\textbf{98.77\%}\\
             &SSIM&0.9475&0.9905&0.9995&\color{black}0.9145&\textbf{0.9996}\\
             &LPIPS&0.1510&0.0457&0.0843&\color{black}0.6124&\textbf{0.0110}\\
             \hline
	    \multirow{4}{*}{\shortstack{GTSRB}}&ASR&96.80\%&97.53\%&82.50\%&\color{black}99.08\%&\textbf{99.30\%}\\
             &CDA&96.07\%&88.03\%&86.80\%&\color{black}\textbf{96.96\%}&96.16\%\\
             &SSIM&0.9473&0.9906&0.9994&\color{black}0.9188&\textbf{0.9995}\\
             &LPIPS&0.1333&0.0479&0.0688&\color{black}0.5340&\textbf{0.0161}\\
             \hline
	    \multirow{4}{*}{\shortstack{CIAFR-100}}&ASR&97.67\%&94.02\%&78.80\%&\color{black}98.96\%&\textbf{99.88\%}\\
             &CDA&91.33\%&\textbf{98.23\%}&82.20\%&\color{black}88.02\%&82.26\%\\
             &SSIM&0.9474&0.9905&0.9993&\color{black}0.9124&\textbf{0.9995}\\
             &LPIPS&0.1489&0.0443&0.0712&\color{black}0.5781&\textbf{0.0117}\\
             \hline
	    \multirow{4}{*}{\shortstack{ImageNette}}&ASR&94.73\%&94.20\%&87.20\%&\color{black}\textbf{96.00\%}&95.08\%\\
             &CDA&81.25\%&88.45\%&84.80\%&\color{black}88.93\%&\textbf{89.63\%}\\
             &SSIM&0.9472&0.9906&\textbf{0.9995}&\color{black}0.9138&\textbf{0.9995}\\
             &LPIPS&0.1281&0.0379&0.0672&\color{black}0.5103&\textbf{0.0124}\\
             \hline
	     \multirow{4}{*}{\shortstack{VGG-Flower-h}}&ASR&92.10\%&96.20\%&77.60\%&\color{black}95.10\%&\textbf{96.51\%}\\
             &CDA&91.10\%&95.45\%&79.20\%&\color{black}\textbf{96.10\%}&95.50\%\\
             &SSIM&0.9455&0.9917&0.9994&\color{black}0.9125&\textbf{0.9995}\\
             &LPIPS&0.1124&0.9911&0.0899&\color{black}0.5989&\textbf{0.0101}\\
		\bottomrule
	\end{tabular}
\end{table*}
\begin{figure*}[tt]
	\centering
	\begin{minipage}[t]{6.7in}
		\centering
		\includegraphics[trim=0mm 0mm 0mm 0mm, clip,width=6.7in]{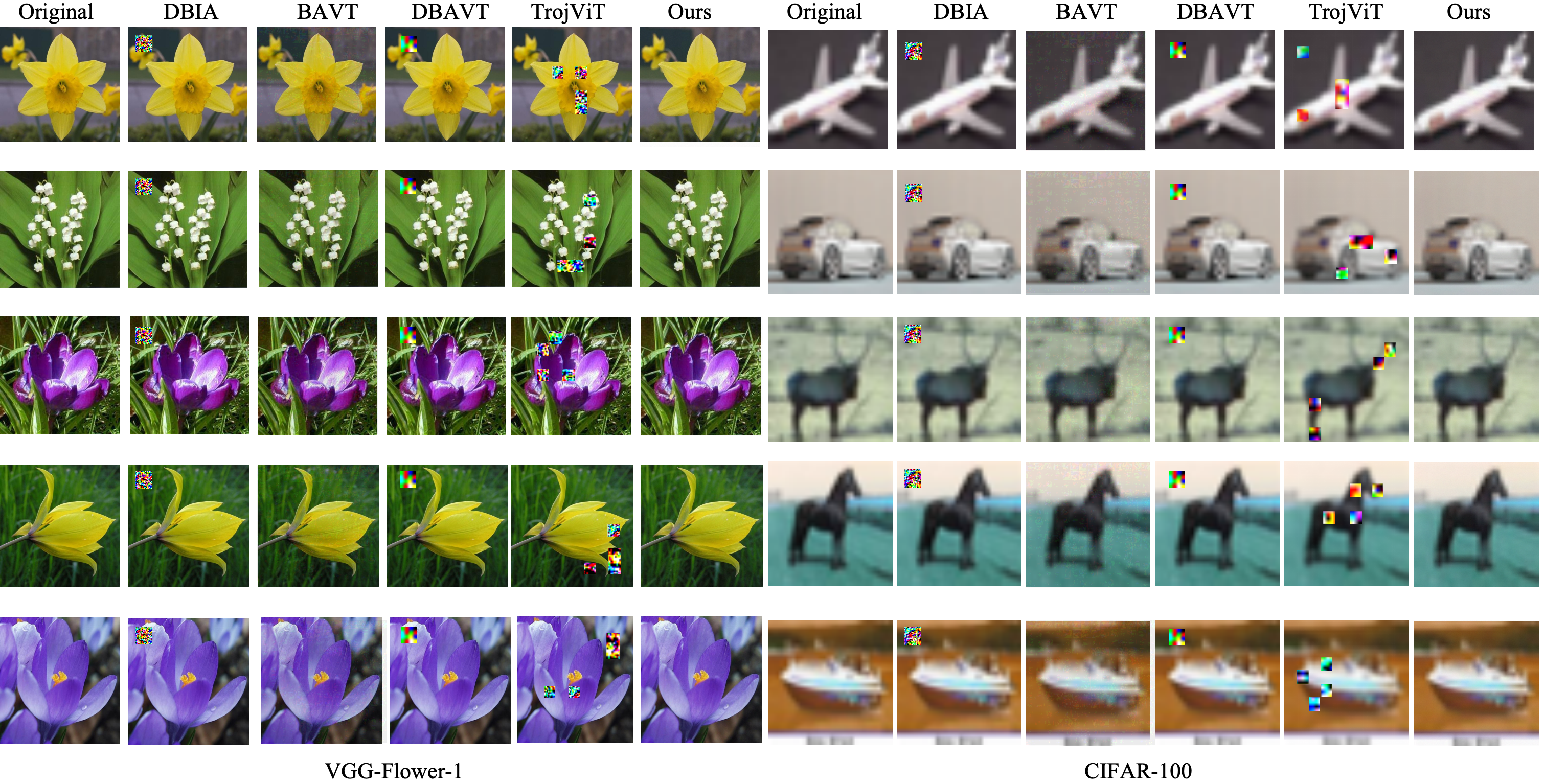}
	\end{minipage}
	\caption{{\color{black}Comparison of backdoored samples between our method and the baselines against ViTs.}} \label{fig:poisoned-vit}
\end{figure*}

\begin{table}[tt]
	\caption{The impact of attention-based mask determination, iterative update, and alternating retraining on our proposed attacks. The target victim models are DNN models.}
	\label{tab:attention2}
	\centering
	\setlength\tabcolsep{3pt}
	\scriptsize
	\begin{tabular}{c|cccccc cc}
		\toprule
		\multirow{3}{*}{\shortstack{Size}}&\multicolumn{8}{c}{VGG-Flower-l}\\&\multicolumn{2}{c}{Base}
		&\multicolumn{2}{c}{Base+Attn} &\multicolumn{2}{c}{Base+Attn+Iter}&\multicolumn{2}{c}{All}\\ &ASR &CDA  & ASR &CDA  & ASR &CDA& ASR &CDA\\
		\midrule
		$1 \times 1$ &18.50\%&93.50\% &29.00\%& 92.50\% & 34.50\% & \textbf{94.50\%} &\textbf{35.00\%} & \textbf{94.50\%}\\
		$2 \times 2$ & 32.00\% & 96.00\% & 42.00\% & 95.50\%& 60.00\% & \textbf{97.00\%}& \textbf{71.50\%} & \textbf{97.00\%}\\
		
		$3 \times 3$ &48.50\% & 93.50\%&74.50\%&94.50\%&\textbf{100.0\%}&95.50\%& 99.50\% & \textbf{96.00\%}\\ 
		$4 \times 4$ &51.00\% & 94.50\%&98.00\%&95.50\%&\textbf{100.0\%}&96.50\%& \textbf{100.0\%} & \textbf{98.00\%}\\ 
		
		\hline
		\multirow{3}{*}{\shortstack{Size}}&\multicolumn{8}{c}{CIFAR-10}\\&\multicolumn{2}{c}{Base}
		&\multicolumn{2}{c}{Base+Attn} &\multicolumn{2}{c}{Base+Attn+Iter}&\multicolumn{2}{c}{All}\\ &ASR &CDA  & ASR &CDA  & ASR &CDA& ASR &CDA\\
		\midrule
	
		$1 \times 1$ &44.89\% &87.22\% & 55.63\% & 86.66\%
		&\textbf{81.33\%} & 87.71\% & 80.33\% & \textbf{87.94\%}\\
		$2 \times 2$ &58.26\% & 87.94\% & 95.60\% & 87.88\%
		&\textbf{99.44\%} & 89.28\% & 99.14\% & \textbf{90.07\%}\\
		$3 \times 3$ &91.01\%  & 87.55\%  & 97.70\%& 88.43\%
		& \textbf{99.62\%}  & 89.07\% & 99.56\% & \textbf{90.23\%}\\ 
		$4 \times 4$ &95.62\%  & 88.39\% & 98.10\% & 88.76\%
		& \textbf{99.77\%}  & 89.35\% & 97.55\% & \textbf{89.91\%}\\ 
		\hline
		\multirow{3}{*}{\shortstack{Size}}&\multicolumn{8}{c}{GTSRB}\\&\multicolumn{2}{c}{Base}
		&\multicolumn{2}{c}{Base+Attn} &\multicolumn{2}{c}{Base+Attn+Iter}&\multicolumn{2}{c}{All}\\  &ASR &CDA  & ASR &CDA  & ASR &CDA& ASR &CDA\\
		\midrule
		$1 \times 1$ & 78.67\% & 93.29\% & 87.67\% & 96.06\% & 97.98\% & 96.67\% &\textbf{98.78\%} & \textbf{97.03\%}\\
		$2 \times 2$ & 75.01\% & 95.52\% & 97.01\% & 95.25\% & \textbf{99.73\%} & 97.14\% &99.00\% & \textbf{97.38\%}\\
		$3 \times 3$ & 93.49\% & 96.75\% & 94.72\% & 96.81\% & 98.97\% & 96.69\% &\textbf{99.98\%} &\textbf{97.00\%}\\ 
		$4 \times 4$ & 91.74\% & 96.89\% & 93.40\% & 97.74\% & 99.80\% & 97.50\%& \textbf{99.87\%} & \textbf{97.78\%}\\ 
	    \hline
		\multirow{3}{*}{\shortstack{Size}}&\multicolumn{8}{c}{CIFAR-100}\\
		&\multicolumn{2}{c}{Base} & \multicolumn{2}{c}{Base+Attn} &\multicolumn{2}{c}{Base+Attn+Iter}&\multicolumn{2}{c}{All}\\  &ASR &CDA  & ASR &CDA  & ASR &CDA& ASR &CDA\\
		\midrule
		
		$1 \times 1$ &93.54\% & 71.84\% &95.41\% & 72.69\%
		& 97.33\% &74.61\%& \textbf{97.61\%} & \textbf{74.66\%}\\
		$2 \times 2$ &97.33\% & 71.67\% &99.56\% & 71.60\%
		&\textbf{99.95\%} & 74.22\% & 99.71\% &\textbf{75.07\%}\\
		$3 \times 3$ &99.39\% & 72.08\% & 99.81\%  & 72.64\%
		&\textbf{99.98\% }& 75.31\%& 99.71\% & \textbf{75.34\%}\\ 
		$4 \times 4$ &99.19\% &72.96\% & 99.28\% &73.76\%
		&\textbf{99.71\%} &73.80\% & 99.64\% & \textbf{75.23\%}\\ 
		\hline
		\multirow{3}{*}{\shortstack{\color{black}Size}}&\multicolumn{8}{c}{ImageNette}\\
		&\multicolumn{2}{c}{\color{black}Base} & \multicolumn{2}{c}{\color{black}Base+Attn} &\multicolumn{2}{c}{\color{black}Base+Attn+Iter}&\multicolumn{2}{c}{\color{black}All}\\  &\color{black}ASR &\color{black}CDA  & \color{black}ASR &\color{black}CDA  & \color{black}ASR &\color{black}CDA& \color{black}ASR &\color{black}CDA\\
		\midrule
		\color{black}$2 \times 2$ &\color{black}51.69\%&\color{black}88.14\%&\color{black}78.62\%&\color{black}88.76\%&\color{black}82.91\%&\color{black}88.73\%&\color{black}\textbf{90.94\%}&\textbf{\color{black}91.26\%}\\
		\color{black}$4 \times 4$ &\color{black}69.83\%&\color{black}82.22\%&\color{black}86.57\%&\color{black}83.39\%&\color{black}89.88\%&\color{black}88.79\%&\color{black}\textbf{90.57\%}&\color{black}\textbf{90.32\%}\\
		\color{black}$8 \times 8$ &\color{black}79.11\%&\color{black}83.54\%&\color{black}88.10\%&\color{black}86.14\%&\color{black}92.51\%&\color{black}86.14\%&\color{black}\textbf{98.39\%}&\color{black}\textbf{90.93\%}\\ 
		\color{black}$12 \times12$ &\color{black}80.05\%&\color{black}82.50\%&\color{black}90.33\%&\color{black}83.18\%&\color{black}92.20\%&\color{black}87.77\%&\color{black}\textbf{99.57\%}&\color{black}\textbf{88.59\%}\\ 
		
		\hline
		\multirow{3}{*}{\shortstack{\color{black}Size}}&\multicolumn{8}{c}{VGG-Flower-h}\\
		&\multicolumn{2}{c}{\color{black}Base} & \multicolumn{2}{c}{\color{black}Base+Attn} &\multicolumn{2}{c}{\color{black}Base+Attn+Iter}&\multicolumn{2}{c}{\color{black}All}\\  &\color{black}ASR &\color{black}CDA  & \color{black}ASR &\color{black}CDA  & \color{black}ASR &\color{black}CDA& \color{black}ASR &\color{black}CDA\\
		\midrule
		\color{black}$8 \times 8$  &\color{black}46.50\%&\color{black}94.50\%&\color{black}69.00\%&\color{black}94.50\%&\color{black}74.50\%&\color{black}94.00\%&\color{black}\textbf{98.50\%}&\color{black}\textbf{97.00\%}\\
		\color{black}$12 \times12$ &\color{black}47.00\%&\color{black}94.50\% &\color{black}77.00\%&\color{black}\textbf{95.50\%}&\color{black}93.00\%&\color{black}\textbf{95.50\%}&
		\color{black}\textbf{98.50\%}&\color{black}\textbf{95.50\%}\\
		\color{black}$16 \times16$ &\color{black}51.00\%&\color{black}95.50\% &\color{black}85.50\%&\color{black}96.50\%&\color{black}94.50\%&\color{black}96.50\%&\color{black}\textbf{99.00\%}&\color{black}\textbf{97.00\%}\\ 
		\color{black}$20 \times20$ &\color{black}56.50\%&\color{black}96.00\% &\color{black}86.00\%&\color{black}95.00\%&\color{black}95.00\%&\color{black}96.00\%&\color{black}\textbf{99.50\%}&\color{black}\textbf{97.00\%}\\ 
		\bottomrule
	\end{tabular}
	
\end{table}

\begin{table*}[tt]
	\caption{The impact of attention-based mask determination, iterative update, and alternating retraining on our proposed attacks. The target victim models are ViT models.}
	\label{tab:abtransformer}
	\centering
	\footnotesize
	\begin{tabular}{c|cc|cc|cc|cc}
		\toprule
		\multirow{2}{*}{\shortstack{Method}}
          &\multicolumn{2}{c}{Base}
		&\multicolumn{2}{c}{Base+Attn} &\multicolumn{2}{c}{Base+Attn+Iter}&\multicolumn{2}{c}{All}\\ & ASR & CDA & ASR & CDA  & ASR & CDA & ASR & CDA\\
		\midrule
		VGG-Flower-l&19.50\% &95.00\%&51.00\%&95.50\% &94.50\% &95.50\% &\textbf{95.70\%}&\textbf{99.00\%}\\
		CIFAR-10&15.47\% & 74.66\%&92.82\% & 80.39\% &\textbf{99.86\%}& 85.56\%&98.89\% & \textbf{89.82\%}\\ 
         GTSRB&12.96\% & 53.19\%&76.25\% & 73.72\%&\textbf{99.89\%}&93.39\% &99.30\% & \textbf{95.32\%}\\
         CIFAR-100&92.75\% & 48.60\%& 99.92\% & 53.09\%&\textbf{99.98\%}& 59.92\%&99.88\% & \textbf{75.75\%}\\
         ImageNette&79.84\% & 83.10\%& 88.87\% & 84.15\%& \textbf{95.65\%}& 87.41\% &95.08\%& \textbf{89.63\%}\\
         VGG-Flower-h&67.50\% & 94.00\% & 75.50\% & 94.50\%  & 96.00\% & \textbf{95.50\%}& \textbf{96.50\%}& \textbf{95.50\%}\\
         
		\bottomrule
	\end{tabular}
\end{table*}

\begin{figure}[tt]
	\centering
	\hspace{0cm}
    \begin{minipage}[t]{1.1in}
		\centering
		\includegraphics[trim=0mm 0mm 0mm 0mm, clip,width=1.1in]{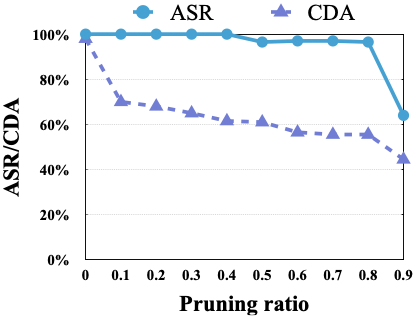}\\
		\centerline{\footnotesize VGG-Flower-l}
	\end{minipage}
	\begin{minipage}[t]{1.1in}
		\centering
		\includegraphics[trim=0mm 0mm 0mm 0mm, clip,width=1.1in]{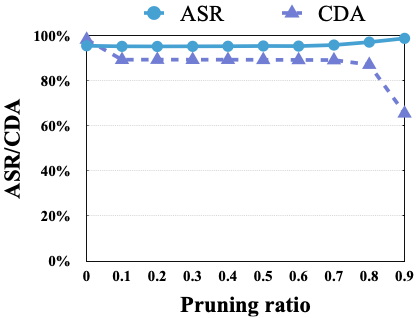}\\
		\centerline{\footnotesize CIFAR-10}
	\end{minipage}
	\begin{minipage}[t]{1.1in}
		\centering
		\includegraphics[trim=0mm 0mm 0mm 0mm, clip,width=1.1in]{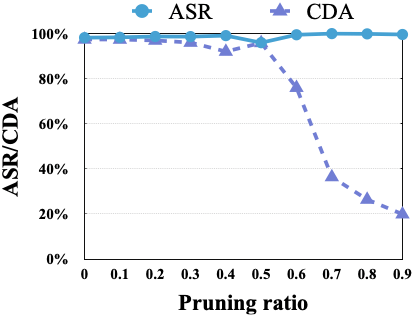}\\
		\centerline{\footnotesize GTSRB}
	\end{minipage}
	\begin{minipage}[t]{1.1in}
		\centering
		\includegraphics[trim=0mm 0mm 0mm 0mm, clip,width=1.1in]{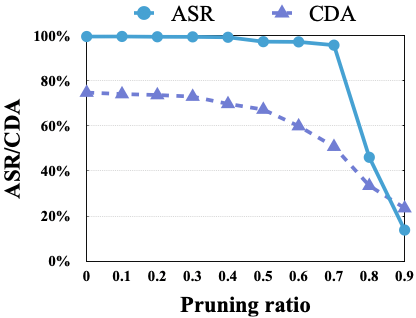}\\
		\centerline{\footnotesize CIFAR-100}
	\end{minipage}
	\begin{minipage}[t]{1.1in}
		\centering
		\includegraphics[trim=0mm 0mm 0mm 0mm, clip,width=1.1in]{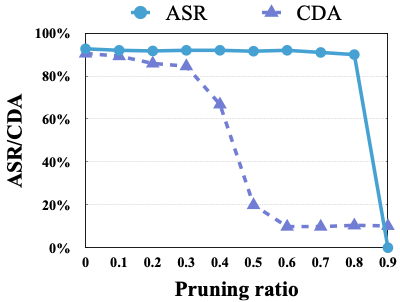}\\
		\centerline{\footnotesize ImageNette}
	\end{minipage}
	\begin{minipage}[t]{1.1in}
		\centering
		\includegraphics[trim=0mm 0mm 0mm 0mm, clip,width=1.1in]{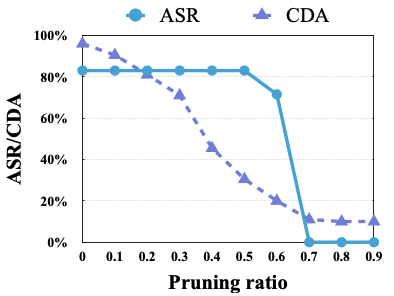}\\
		\centerline{\footnotesize VGG-Flower-h}
	\end{minipage}
	\caption{{\color{black}The attack performance after applying model pruning to our proposed attack.  }} \label{fig:prune}
\end{figure}
\begin{figure}[tt]
	\centering
	\begin{minipage}[t]{1.1in}
		\centering
		\includegraphics[trim=0mm 0mm 0mm 0mm, clip,width=1.1in]{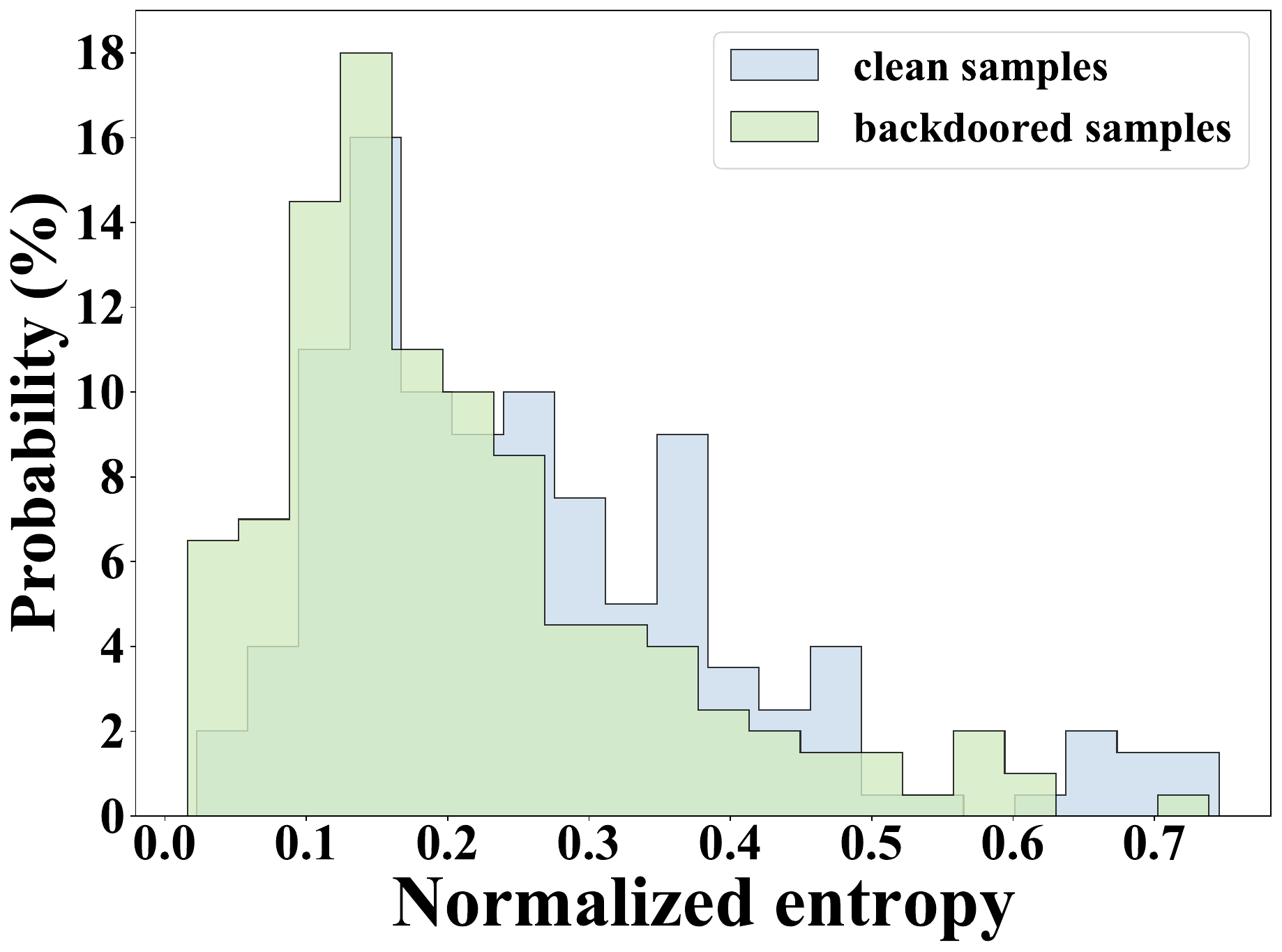}\\
		\centerline{\footnotesize VGG-Flower-l}
	\end{minipage}
	\begin{minipage}[t]{1.1in}
		\centering
		\includegraphics[trim=0mm 0mm 0mm 0mm, clip,width=1.1in]{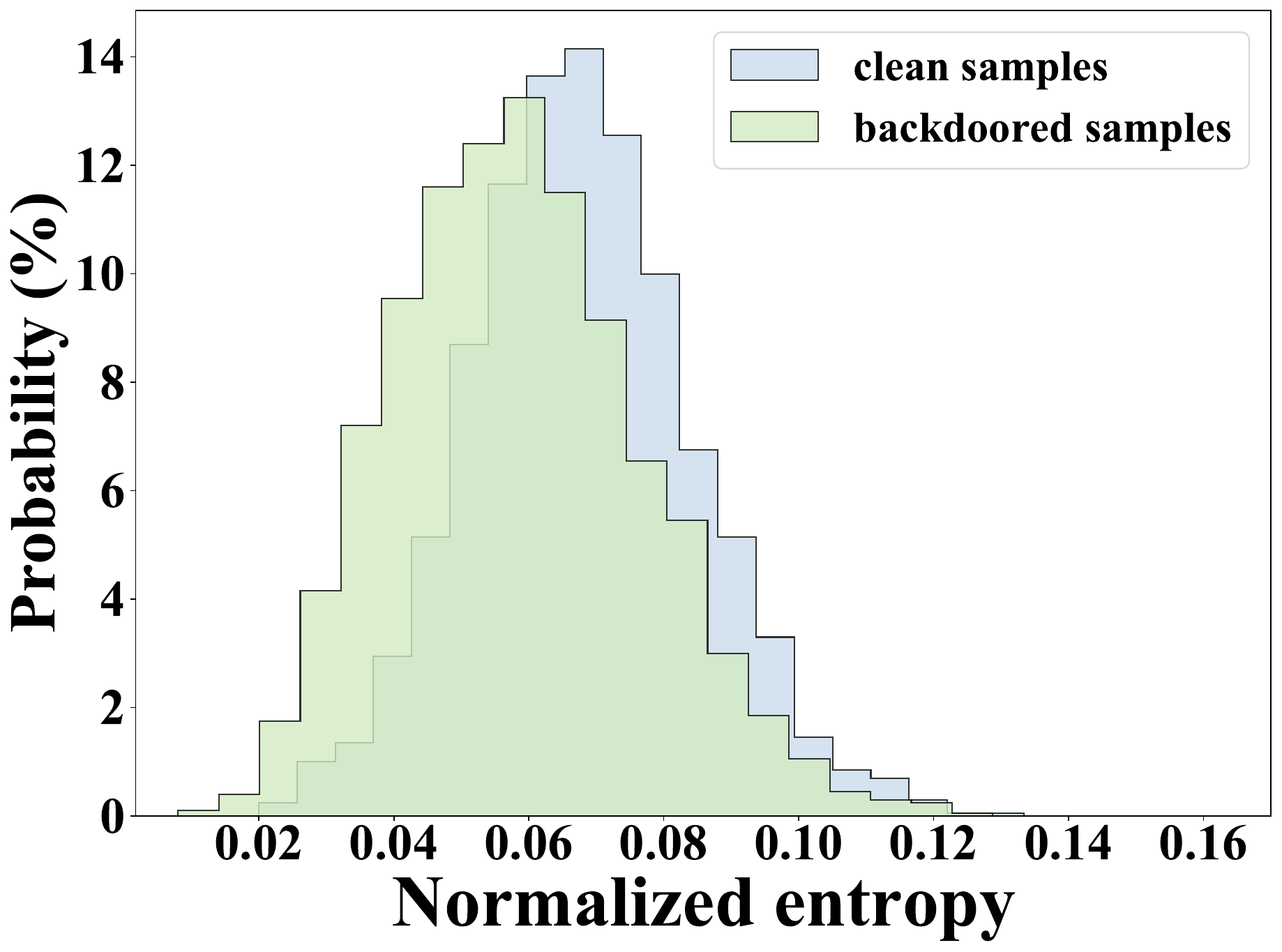}\\
		\centerline{\footnotesize CIFAR-10}
	\end{minipage}
	\begin{minipage}[t]{1.1in}
		\centering
		\includegraphics[trim=0mm 0mm 0mm 0mm, clip,width=1.1in]{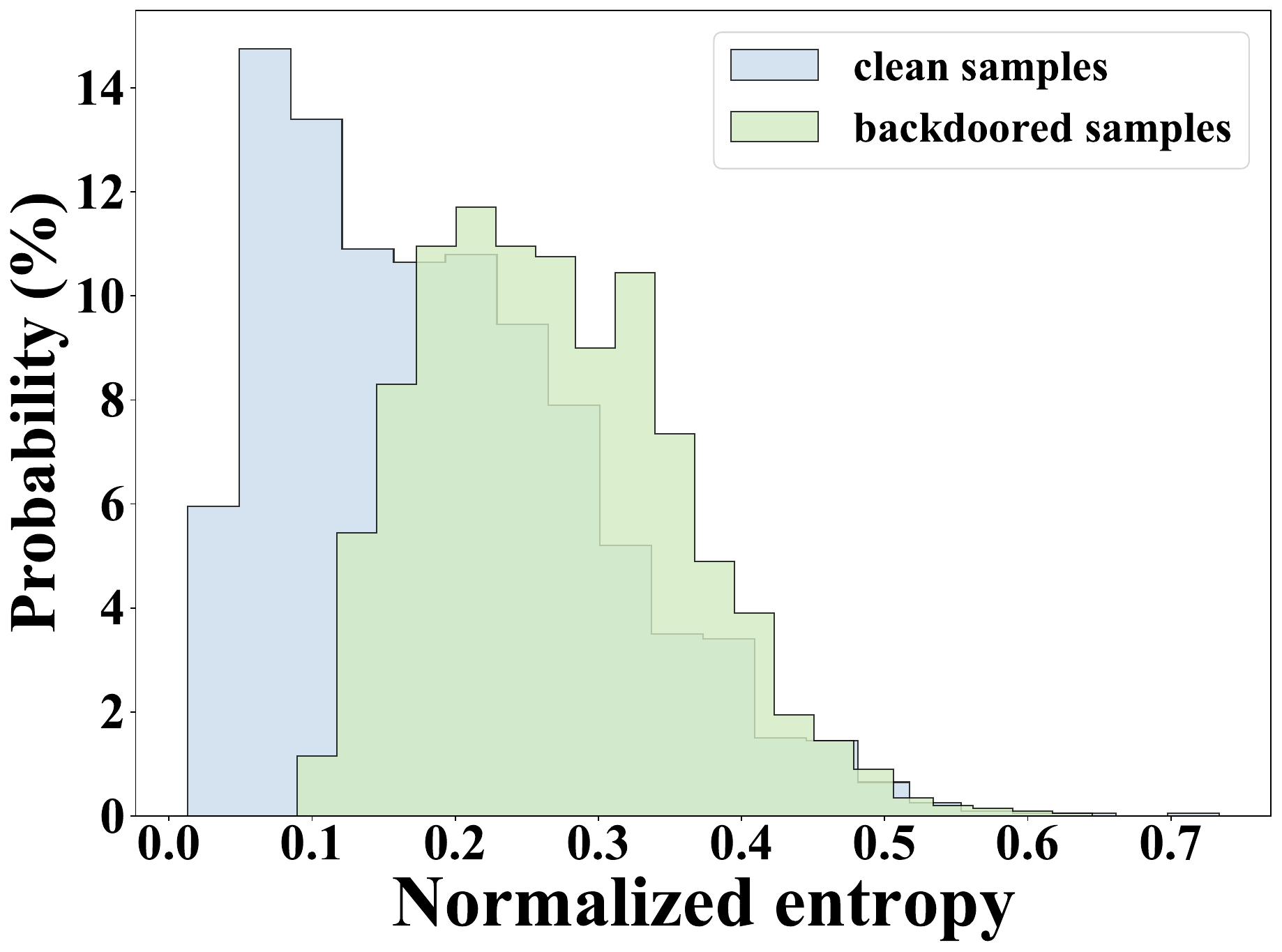}\\
		\centerline{\footnotesize GTSRB}
	\end{minipage}
	\begin{minipage}[t]{1.1in}
		\centering
		\includegraphics[trim=0mm 0mm 0mm 0mm, clip,width=1.1in]{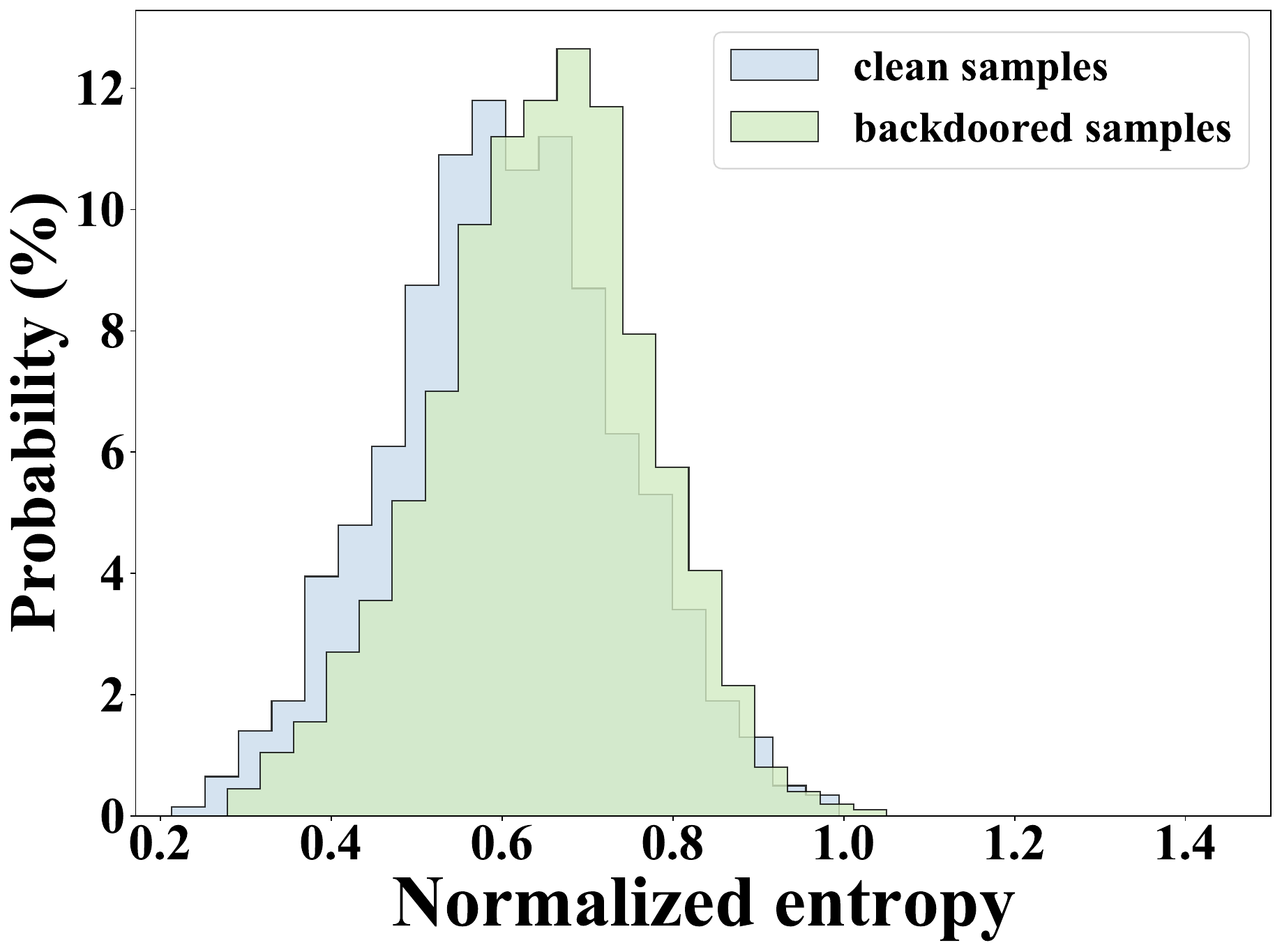}\\
		\centerline{\footnotesize CIFAR-100 }
	\end{minipage}
	\begin{minipage}[t]{1.1in}
		\centering
		\includegraphics[trim=0mm 0mm 0mm 0mm, clip,width=1.1in]{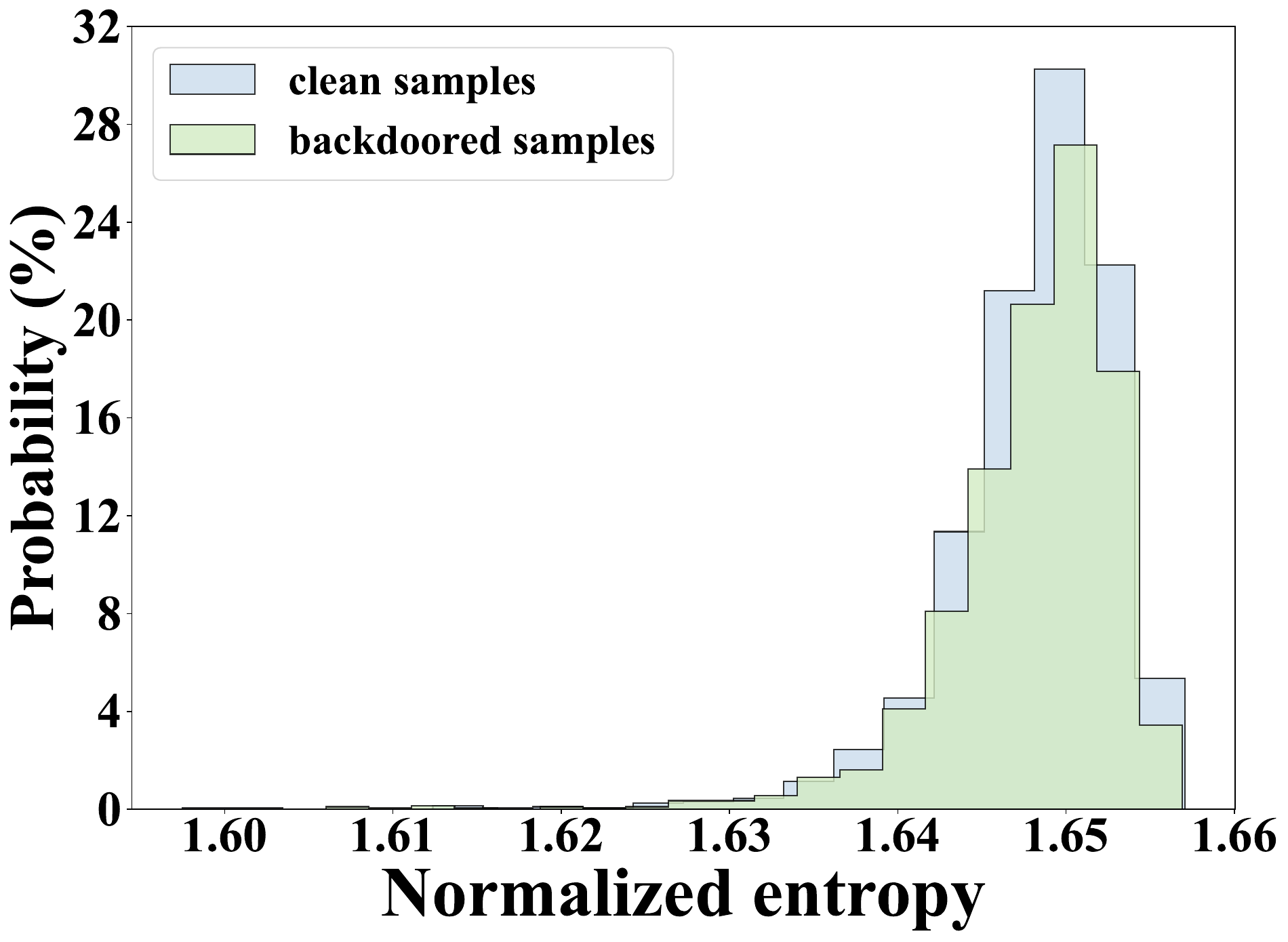}\\
		\centerline{\footnotesize ImageNette}
	\end{minipage}
	\begin{minipage}[t]{1.1in}
		\centering
		\includegraphics[trim=0mm 0mm 0mm 0mm, clip,width=1.1in]{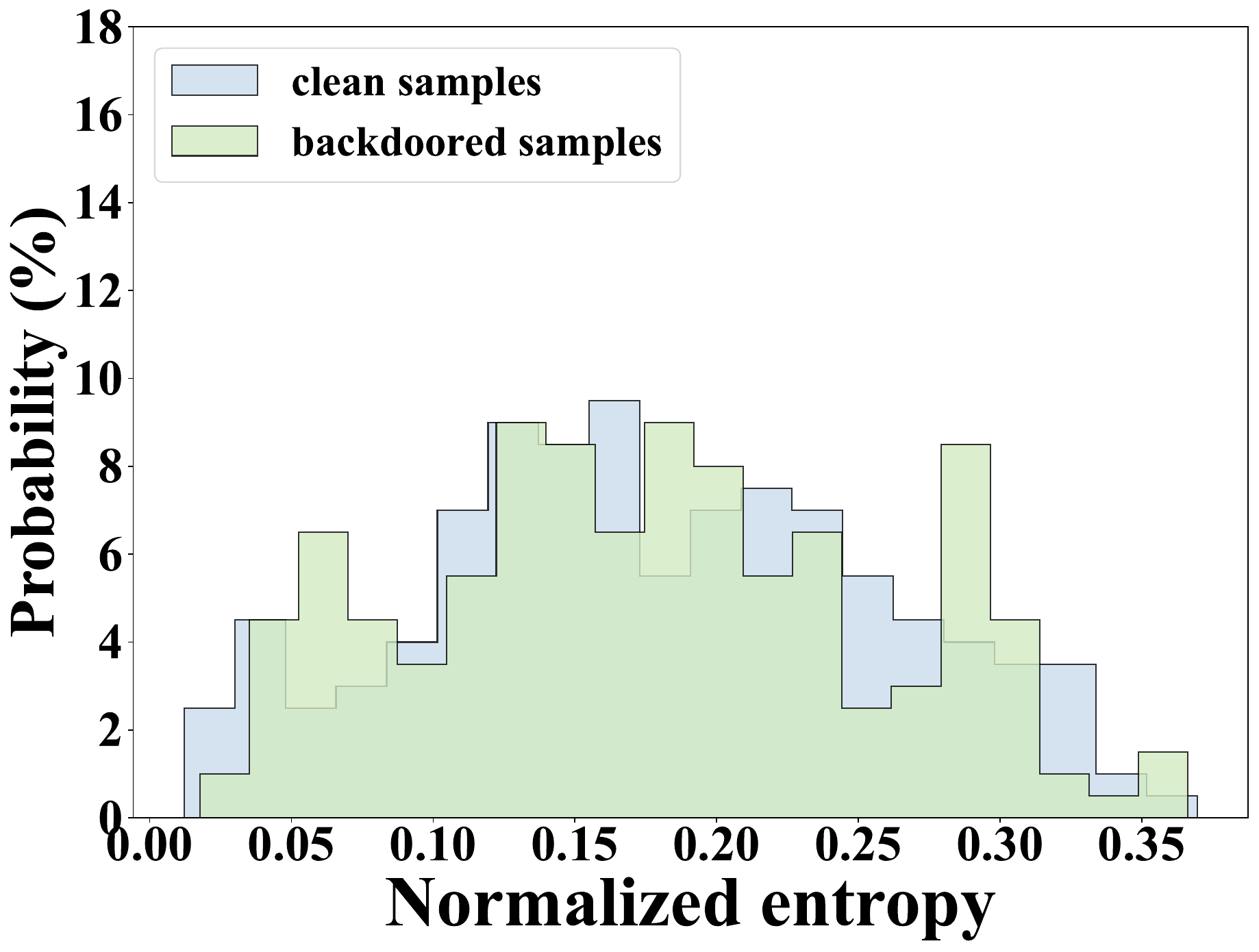}\\
		\centerline{\footnotesize VGG-Flower-h}
	\end{minipage}
	\caption{{\color{black}The distribution of the entropy prediction results of clean samples and backdoored samples after applying STRIP to our proposed attack. }} \label{fig:STRIP}
\end{figure}

\begin{figure}[t]
    \centering
    \begin{minipage}[t]{1.1in}
        \centering
        \includegraphics[trim=0mm 0mm 0mm 0mm, clip,width=0.45in]{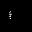}
        \includegraphics[trim=0mm 0mm 0mm 0mm, clip,width=0.45in]{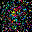}
        \centerline{\footnotesize VGG-Flower-l}
    \end{minipage}
    \begin{minipage}[t]{1.1in}
        \centering
        \includegraphics[trim=0mm 0mm 0mm 0mm, clip,width=0.45in]{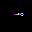}
        \includegraphics[trim=0mm 0mm 0mm 0mm, clip,width=0.45in]{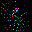}
        \centerline{\footnotesize CIFAR-10}
    \end{minipage}
    \begin{minipage}[t]{1.1in}
        \centering
        \includegraphics[trim=0mm 0mm 0mm 0mm, clip,width=0.45in]{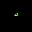}
        \includegraphics[trim=0mm 0mm 0mm 0mm, clip,width=0.45in]{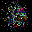}
        \centerline{\footnotesize GTSRB}
    \end{minipage}
    \\ \vspace{0.02\linewidth}
    \begin{minipage}[t]{1.1in}
        \centering
\includegraphics[trim=0mm 0mm 0mm 0mm, clip, width=0.45in]{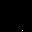}
\includegraphics[trim=0mm 0mm 0mm 0mm, clip, width=0.45in]{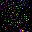}

        \centerline{\footnotesize CIFAR-100}
    \end{minipage}
    \begin{minipage}[t]{1.1in}
        \centering
        \includegraphics[trim=0mm 0mm 0mm 0mm, clip,width=0.45in]{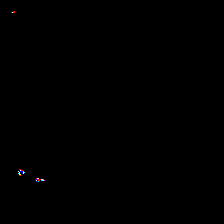}
        \includegraphics[trim=0mm 0mm 0mm 0mm, clip,width=0.45in]{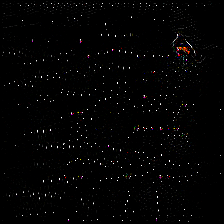}
        \centerline{\footnotesize ImageNette}
    \end{minipage}
    \begin{minipage}[t]{1.1in}
        \centering
        \includegraphics[trim=0mm 0mm 0mm 0mm, clip,width=0.45in]{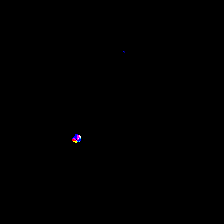}
        \includegraphics[trim=0mm 0mm 0mm 0mm, clip,width=0.45in]{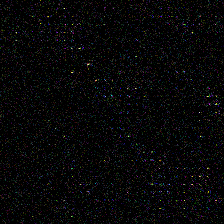}
        \centerline{\footnotesize VGG-Flower-h}
    \end{minipage}
    \caption{{\color{black}The comparison between the actual triggers and the triggers recovered by NC for various attacks. In each pair, the left image depicts the real trigger, while the right image shows the recovered trigger.}} \label{fig:NC}
\end{figure}

\begin{table}[tt]
	\caption{{\color{black}The impact of key neuron gradient boosting on attack performance. In this case, the target victim models are DNN models. }}
	\label{tab:neuron-dnn}
	\centering
	\footnotesize
	\begin{tabular}{>{\color{black}}c|>{\color{black}}l>{\color{black}}c>{\color{black}}c>{\color{black}}c>{\color{black}}c>{\color{black}}c>{\color{black}}c>{\color{black}}c>{\color{black}}c>{\color{black}}c>{\color{black}}c>{\color{black}}c>{\color{black}}c}
		\toprule
		 \multirow{2}{*}{Datasets}
		&\multicolumn{2}{c}{\color{black}Without gradient boosting} &\multicolumn{2}{c}{\color{black}With gradient boosting} \\ &ASR &CDA& ASR &CDA \\
		\midrule
		\multirow{1}{*}{\shortstack{VGG-Flower-l}}&98.10\%&91.70\%&\textbf{99.50\%}&\textbf{96.00\%}\\
	    \multirow{1}{*}{\shortstack{CIFAR-10}}&89.10\%&88.25\%&\textbf{99.56\%}&\textbf{90.23\%}\\
	    \multirow{1}{*}{\shortstack{GTSRB}}&92.12\%&94.28\%&\textbf{99.98\%}&\textbf{97.00\%}\\
	    \multirow{1}{*}{\shortstack{CIAFR-100}}&99.31\%&73.10\%&\textbf{99.71\%}&\textbf{75.34\%}\\
	    \multirow{1}{*}{\shortstack{ImageNette}}&91.72\%&78.97\%&\textbf{98.39\%}&\textbf{90.93\%}\\
	    \multirow{1}{*}{\shortstack{VGG-Flower-h}}&86.00\%&92.00\%&\textbf{99.00\%}&\textbf{97.00\%}\\
		
		\bottomrule
	\end{tabular}
	
\end{table}
\begin{table}[tt]
	\caption{{\color{black}The impact of key neuron gradient boosting on attack performance. In this case, the target victim models are ViT models. }}
	\label{tab:neuron-vit}
	\centering
	\footnotesize
	\begin{tabular}{>{\color{black}}c|>{\color{black}}l>{\color{black}}c>{\color{black}}c>{\color{black}}c>{\color{black}}c>{\color{black}}c>{\color{black}}c>{\color{black}}c>{\color{black}}c>{\color{black}}c>{\color{black}}c>{\color{black}}c>{\color{black}}c}
		\toprule
		 \multirow{2}{*}{Datasets}
		&\multicolumn{2}{c}{\color{black}Without gradient boosting} &\multicolumn{2}{c}{\color{black}With gradient boosting} \\ &ASR &CDA& ASR &CDA \\
		\midrule
		\multirow{1}{*}{\shortstack{VGG-Flower-l}}&97.70\%&83.50\%&\textbf{98.80\%}&\textbf{96.40\%}\\
	    \multirow{1}{*}{\shortstack{CIFAR-10}}&88.36\%&79.41\% &\textbf{99.86\%}&\textbf{89.74\%}\\
	    \multirow{1}{*}{\shortstack{GTSRB}}&82.95\%&94.79\%&\textbf{99.27\%}&\textbf{95.32\%}\\
	    \multirow{1}{*}{\shortstack{CIAFR-100}}&\textbf{99.92\%}&72.43\%&99.32\%&\textbf{74.99\%}\\
	    \multirow{1}{*}{\shortstack{ImageNette}}&88.12\%&76.97\%&\textbf{90.34\%}&\textbf{87.41\%}\\
	    \multirow{1}{*}{\shortstack{VGG-Flower-h}}&82.00\%&76.00\%&\textbf{84.00\%}&\textbf{94.00\%}\\
		
		\bottomrule
	\end{tabular}
	
\end{table}

\begin{table}[h]
	\caption{{\color{black}Impact of the neuron residing layer.}}
	\label{tab:residing layer}
	\centering
	\begin{tabular}{ll|cc}
		\toprule
		\color{black}Dataset&\color{black}Selected layer& \color{black}ASR &\color{black}CDA\\
		\hline
		\multirow{4}{*}{\color{black}VGG-Flower-l}  &\color{black}Patch embedding&\color{black}95.05\%\color{black}&\color{black}97.97\%\\ &\color{black}Attention layer 2&\color{black}95.67\%&\color{black}98.65\%\\
		&\color{black}Attention layer 5&\color{black}95.93\%&\color{black}98.08\%\\
		&\color{black}Attention layer 8&\color{black}97.98\%&\color{black}97.04\%\\
        &\color{black}Attention layer 11&\color{black}98.57\%&\color{black}99.05\%\\
		&\color{black}Head layer&\color{black}98.70\%&\color{black}99.00\%\\
		\hline
		\multirow{4}{*}{\color{black}CIFAR-10}  &\color{black}Patch embedding&\color{black}89.01\%&\color{black}89.90\%\\
       &\color{black}Attention layer 2&\color{black}89.03\%&\color{black}89.12\%\\
		&\color{black}Attention layer 5&\color{black}92.07\%&\color{black}90.10\%\\
		&\color{black}Attention layer 8&\color{black}98.03\%&\color{black}90.13\%\\
        &\color{black}Attention layer 11&\color{black}98.08\%&\color{black}89.28\%\\
		&\color{black}Head layer&\color{black}98.89\%&\color{black}89.92\%\\
		\hline
		\multirow{4}{*}{\color{black}{GTSRB}}  &\color{black}Patch embedding&\color{black}92.35\%&\color{black}95.90\%\\
    &\color{black}Attention layer 2&\color{black}92.93\%&\color{black}96.47\%\\
		&\color{black}Attention layer 5&\color{black}93.00\%&\color{black}96.03\%\\
		&\color{black}Attention layer 8&\color{black}98.08\%&\color{black}95.70\%\\
        &\color{black}Attention layer 11&\color{black}98.80\%&\color{black}97.37\%\\
		&\color{black}Head layer&\color{black}99.30\%&\color{black}95.32\%\\
		\hline
		\multirow{4}{*}{\color{black}{CIFAR-100}}  &\color{black}Patch embedding&\color{black}94.02\%&\color{black}75.87\%\\
    &\color{black}Attention layer 2&\color{black}94.08\%&\color{black}75.03\%\\
		&\color{black}Attention layer 5&\color{black}94.95\%&\color{black}74.89\%\\
		&\color{black}Attention layer 8&\color{black}99.07\%&\color{black}75.99\%\\
        &\color{black}Attention layer 11&\color{black}99.07\%&\color{black}75.68\%\\
		&\color{black}Head layer&\color{black}99.88\%&\color{black}75.75\%\\
		\hline
		\multirow{4}{*}{\color{black}ImageNette}  &\color{black}Patch embedding&\color{black}86.94\%&\color{black}88.25\%\\
  &\color{black}Attention layer 2&\color{black}87.08\%&\color{black}87.63\%\\
		&\color{black}Attention layer 5&\color{black}87.32\%&\color{black}86.80\%\\
		&\color{black}Attention layer 8&\color{black}87.84\%&\color{black}88.98\%\\
        &\color{black}Attention layer 11&\color{black}91.07\%&\color{black}89.84\%\\
		&\color{black}Head layer&\color{black}91.08\%&\color{black}89.63\%\\
            \hline
		\multirow{4}{*}{\color{black}VGG-Flower-h}  &\color{black}Patch embedding&\color{black}82.00\%&\color{black}94.00\%\\
  &\color{black}Attention layer 2&\color{black}82.00\%&\color{black}95.00\%\\
		&\color{black}Attention layer 5&\color{black}82.50\%&\color{black}95.00\%\\
		&\color{black}Attention layer 8&\color{black}84.00\%&\color{black}96.00\%\\
        &\color{black}Attention layer 11&\color{black}84.00\%&\color{black}95.50\%\\
		&\color{black}Head layer&\color{black}84.50\%&\color{black}95.50\%\\
		\bottomrule
	\end{tabular}
\end{table}
\begin{table}[tt]
	\caption{{\color{black}Apply NAD to our proposed method. }}
	\label{tab:nad}
	\centering
	\footnotesize
	\begin{tabular}{>{\color{black}}c|>{\color{black}}l>{\color{black}}c>{\color{black}}c>{\color{black}}c>{\color{black}}c>{\color{black}}c>{\color{black}}c>{\color{black}}c>{\color{black}}c>{\color{black}}c>{\color{black}}c>{\color{black}}c>{\color{black}}c}
		\toprule
		 \multirow{2}{*}{Datasets}
		&\multicolumn{2}{c}{\color{black}Original} &\multicolumn{2}{c}{\color{black}NAD} \\ &ASR &CDA& ASR &CDA \\
		\midrule
		\multirow{1}{*}{\shortstack{VGG-Flower-l}}&99.50\%&97.50\%&92.50\%&97.00\%\\
	    \multirow{1}{*}{\shortstack{CIFAR-10}}&99.76\%&89.46\% &99.19\%&88.31\%\\
	    \multirow{1}{*}{\shortstack{GTSRB}}&99.75\%&97.17\%&90.14\%&96.69\%\\
	    \multirow{1}{*}{\shortstack{CIAFR-100}}&99.58\%&74.62\%&94.23\%&73.92\%\\
	    \multirow{1}{*}{\shortstack{ImageNette}}&92.16\%&92.40\%&90.56\%&92.31\%\\
	    \multirow{1}{*}{\shortstack{VGG-Flower-h}}&83.00\%&96.00\%&80.00\%&94.00\%\\
		
		\bottomrule
	\end{tabular}
	\vspace{-0.4cm}
\end{table}

\begin{table*}[tt]
	\caption{{\color{black}Apply DBAVT defense to our proposed attack and baseline attacks.}}
	\label{tab:dbavt}
	\centering
	\footnotesize
	\begin{tabular}{>{\color{black}}c|>{\color{black}}l>{\color{black}}c>{\color{black}}c>{\color{black}}c>{\color{black}}c>{\color{black}}c>{\color{black}}c>{\color{black}}c>{\color{black}}c>{\color{black}}c>{\color{black}}c>{\color{black}}c>{\color{black}}c}
		\toprule
		 \multirow{2}{*}{Datasets}
		&\multicolumn{2}{c}{\color{black}Original} &\multicolumn{2}{c}{\color{black}DBIA}&\multicolumn{2}{c}{\color{black}DBAVT}&\multicolumn{2}{c}{\color{black}BAVT}&\multicolumn{2}{c}{\color{black}TrojViT}&\multicolumn{2}{c}{\color{black}Ours} \\ &ASR &CDA& ASR &CDA &ASR &CDA& ASR &CDA&ASR &CDA& ASR &CDA\\
		\midrule
		\multirow{1}{*}{\shortstack{VGG-Flower-l}}&98.80\%&96.40\%&57.20\%&96.30\%&45.30\%&94.10\%&72.40\%&95.20\%&58.50\%&93.90\%&73.50\%&96.50\%\\
	    \multirow{1}{*}{\shortstack{CIFAR-10}}&99.86\%&89.74\%&60.81\%&85.10\%&32.00\%&87.00\%&83.03\%&80.62\%&49.33\%&81.02\%&83.47\%&78.98\%\\
	    \multirow{1}{*}{\shortstack{GTSRB}}&99.27\%&95.32\%&58.18\%&94.03\%&48.90\%&91.16\%&89.90\%&92.98\%&40.87\%&91.03\%&90.67\%&90.95\%\\
	    \multirow{1}{*}{\shortstack{CIAFR-100}}&99.32\%&74.99\%&50.98\%&70.03\%&27.01\%&72.39\%&90.10\%&72.39\%&33.87\%&70.71\%&89.73\%&71.32\%\\
	    \multirow{1}{*}{\shortstack{ImageNette}}&90.34\%&87.41\%&59.42\%&85.03\%&41.89\%&86.40\%&89.40\%&86.00\%&47.88\%&86.50\%&88.45\%&86.16\%\\
	    \multirow{1}{*}{\shortstack{VGG-Flower-h}}&84.50\%&94.00\%&50.50\%&91.50\%&28.00\%&91.50\%&70.50\%&89.50\%&39.50\%&90.00\%&72.00\%&90.00\%\\
		
		\bottomrule
	\end{tabular}
	
\end{table*}

\subsection{QoE-based Trigger Generation}\label{sec:4}

\textbf{{\color{black}Neuron selection.}}
{\color{black}Given the trigger mask, the process of trigger generation is equal to seeking the optimal value assignments in the mask. The idea of trigger generation is to find a neuron in the clean model as a bridge between the input trigger and the target output. To find the neuron, we first determine the proper layer at which the neuron should reside and then pinpoint the specific neuron. As for the DNN model, following \cite{Gong2021}, we select the first fully-connected layer and choose the neuron that has the highest number of activations when the model takes a set of clean samples of the target label.

When considering the ViT model, we also choose a neuron that has the highest correlation with the target label. However, due to the inherent differences between the DNN model and the ViT structure, we cannot directly select the first fully-connected layer as the neuron-residing layer. 

The transformer model primarily consists of three components: patch embedding, attention blocks, and a head. Patch embedding converts each input patch into a QKV matrix. The attention blocks employ equation~(\ref{eq:1}) to compute the QKV matrix, incorporating residual connections. The head comprises a fully connected layer, which extracts classification information from the output of the attention blocks. In contrast, CNNs have several fully-connected layers interspersed, while the main structure of the ViT model (attention blocks) primarily involves attention and residual connection operations, without any interspersed fully connected layers. This structural disparity necessitates the reselection of the layer where the key neurons are located.

We discovered that within the patch embedding and attention blocks structure, altering the input of a neuron does not impact the output of all neurons. In contrast, in the head structure, which consists of a fully connected layer, every input is connected to each output with weighted connections. The neuron in the head layer responds strongly to the input trigger and the output results. Therefore, we opt to select the neuron within the head layer. After determining the neuron-residing layer, we also choose the neuron with the highest number of activations when the model takes a set of clean samples of the target label.}

\textbf{{\color{black}QoE-based Trigger Generation.}}
When generating the trigger, we incorporate gradient enhancement techniques for the selected neurons to further enhance the attack effectiveness. During the gradient descent optimization process of the trigger, we assign greater weight to the gradients of the selected neurons. By prioritizing these key neurons, which play a vital role in classifying the target label, we can effectively amplify the poisoning effect of the generated trigger.

Specifically, the optimization process for trigger gradient descent can be described as follows:
\begin{equation}\label{equ:grad}
\begin{aligned}
\mathcal{L} &= \mathcal{L}(x_t, F_A) + \lambda \mathcal{L}_{\delta}(x_t, x) + \eta \cdot SSIM, \\
&T^{i+1} = T^{i} - lr \cdot \nabla_{T^i} M, \\
&\text{s.t. } \nabla_e := \theta \nabla_e,
\end{aligned}
\end{equation}

where $e$ represents the selected neuron(s), $T^i$ is the trigger for the $i$-th round, $\nabla_{T^i}$ and $\nabla_e$ denote the gradients of $T^i$ and $e$ respectively, back-propagated from the loss function $\mathcal{L}$.
$M$ is the mask generated by RAN, and $\theta$ is the augmentation factor, which is 4 for CIFAR-100, 3 for CIFAR-10, 21 for GTSRB, 30 for VGG-Flower-l, 2 for ImageNette, and 30 for VGG-Flower-h. Note that we set the values of different augmentation factors according to the experimental effect.

An invisible backdoor trigger is also the key to a successful backdoor attack. A visible backdoor trigger can be easily detected by human visual inspection.
In this paper, we propose to introduce Structural Similarity Index Measure (SSIM) \cite{wang2004image} to the loss function and adjust the transparency of the backdoor trigger. SSIM is a commonly used Quality-of-Experience (QoE) metric \cite{chen2014qos}) that is used to compare the differences in luminance, contrast, and structure between the original image and the distorted image.
\begin{equation}
SSIM = A(x,x')^{\alpha} B(x,x')^{\beta} C(x,x')^{\gamma},
\end{equation}
where $A(x,x'), B(x,x'), C(x,x')$ quantify the luminance similarity, contrast similarity, and structure similarity between the original image $x$ and the distorted image $x'$. $\alpha, \beta, \gamma$ are parameters. 
We introduce SSIM into the loss function to optimize the trigger.
\begin{equation}
    \delta^*= \arg\min_{\delta} \big(\mathcal{L}(x_t, F_A) + \lambda \mathcal{L}_{\delta} (x_t,x) + \eta    SSIM \big),
\end{equation}
where $\eta$ balances the attack success rate and the QoE of poisoned images. {\color{black}According to our extensive experiments, we empirically set $\eta$ as 0.1.}

To improve the invisibility of the backdoored samples, we also carefully adjusted the transparency of the backdoor trigger. If we use a higher transparency value, the trigger will be more stealthy but making it more challenging to trigger malicious behaviors. Setting a proper transparency value is a trade-off between the attack success rate and the concealment. Through experiments, we set the transparency value as 0.4 (VGG-Flower-l, CIFAR-10, GTSRB, and CIFAR-100) or 0.7 (ImageNette and VGG-Flower-h) by default.

\subsection{Alternating Retraining}

{\color{black}In backdoor attacks, the conventional method to maintain high prediction accuracy involves retraining deep neural networks using pairs of backdoored samples \( x + \delta \) with target label \( t \) and benign samples \( x \) with ground-truth label \( y \). This approach teaches the model to recognize backdoor triggers while retaining accuracy on benign samples. However, we observed that such methods can lead to reduced accuracy on clean data.

To address this issue and make the backdoored model more similar to the benign model, we propose an alternating retraining strategy. In this method, during iterative updates, we retrain the backdoored model using mixed poisoned datasets when the iteration index \( k \) is even, and only benign samples with their true labels when \( k \) is odd. The benefits of this alternating retraining method are twofold. Firstly, it maintains the model's sensitivity to backdoor triggers while preserving its ability to generalize from clean inputs. By intermittently integrating clean samples into training, the model avoids becoming overly specialized to the poisoned samples, thereby enhancing its overall prediction accuracy. Secondly, this method significantly mitigates the risk of overfitting to the specific features of the poisoned data. Regular retraining on benign samples encourages the model to develop more robust feature representation abilities.

Furthermore, we found that this alternating retraining strategy can also help evade certain backdoor defenses, such as MNTD \cite{xu2019detecting}. We attribute it to the fact that the alternating retraining strategy can minimize the difference between the backdoor modeled and the benign one. The details are shown in the experiment results.}

\section{Evaluation Setup}\label{sec:evaluation}
\subsection{Victim Networks}
In this paper, we conduct experiments on various machine learning tasks, covering different datasets (VGG-Flower \cite{nilsback2008102}, CIFAR-10 \cite{krizhevsky2009learning}, GTSRB \cite{StallkampSSI12}, CIFAR-100 \cite{krizhevsky2009learning}, and ImageNette \cite{imagentte}) and deep neural networks. Note that we randomly select 10 classes with 1,673 training images and 200 test images for VGG-Flower. For VGG-Flower-l, the selected images are uniformly resized to 32 $\times$ 32. For VGG-Flower-h, the selected images are uniformly resized to 224 $\times$ 224.
We utilize VGG-16, ResNet-18, VGG-16, ResNet-34, ResNet-50, and ResNet-18 structures to train DNN models for these six datasets, respectively. {\color{black}We employ the ViT model \cite{dosovitskiy2020image} to train ViT models for the six datasets.}

The default target label is label 0 for VGG-Flower-l, label 3 for VGG-Flower-h, label 2 for CIFAR-10, label 10 for GTSRB, label 0 for CIFAR-100, and label 3 for ImageNette. {\color{black}The default poison ratio is 20\% for VGG-Flower-l, 15\% for VGG-Flower-h, 5\% for CIFAR-10, 5\% for GTSRB, 0.5\% for CIFAR-100, and 15\% for ImageNette. The default trigger size is $4 \times 4$ for VGG-Flower-l, $8 \times 8$ for VGG-Flower-h, $4 \times 4$ for CIFAR-10, $3 \times 3$ for GTSRB,  $2 \times 2$ for CIFAR-100, and $8 \times 8$ for ImageNette.}
The default transparency value is 0.4 for VGG-Flower-l, CIFAR-10, GTSRB, CIFAR-100, and 0.7 for ImageNette and VGG-Flower-h. We adopt a 92-layer RAN with 6 attention modules. We set $C_1=128, C_2=256, C_3=256$ following the original RAN model \cite{wang2017residual}, and $C_4 =C_5=C_6=1$ to aggregate all information into a single attention map. {\color{black}As the ViT model requires an input image size of $3 \times 224 \times 224$, this might not be directly suitable for low-resolution images. To overcome this limitation, we preprocess the low-resolution dataset by applying bilinear interpolation to expand the images to a format compatible with the transformer's input requirements.}
The victim DNN model prediction accuracies of these six datasets are 98.5\%, 91.94\%, 97.25\%, 79.09\%, 92.43\%, and 97.5\%, respectively. {\color{black}The victim ViT model prediction accuracy of these six datasets are 99\%, 89.82\%, 95.32\%, 75.75\%, 89.63\%, and 95.5\%, respectively.} {\color{black}Note that the baselines and our proposed method have the same experiment settings (e.g., trigger size, poison ratio, epoch, learning rate) in the attack performance comparison.}

\subsection{Evaluation Metrics}
We utilize ASR, CDA, SSIM, and LPIPS as our evaluation metrics.

ASR measures the effectiveness of the backdoor attacks, computed as the probability that a trigger-imposed sample is misclassified to the target label.

CDA measures whether the backdoored model can maintain the prediction accuracy of clean input samples.

{\color{black}SSIM \cite{chen2014qos} is a widely-used Quality-of-Experience (QoE) metric that measures the differences in luminance, contrast, and structure between an original image and a distorted image. The SSIM value falls within the range of $[0, 1]$, where a higher SSIM indicates a greater similarity between the original and backdoored images.

LPIPS \cite{zhang2018unreasonable} is a metric that quantifies the similarity between two images by leveraging the hierarchical processing of the human visual system. It operates on the notion that lower-level image features, such as edges and textures, are processed before higher-level features like objects and scenes. The LPIPS metric employs a deep neural network to compute the similarity between the two images. LPIPS has demonstrated superior performance compared to other metrics like SSIM in measuring perceptual similarity between images, particularly when the differences lie in high-level perceptual qualities such as texture and style. A smaller LPIPS value indicates a higher degree of similarity between the two images.}




\section{Evaluation Results}



\subsection{Comparison with Baselines against DNN models}

As shown in Table~\ref{tab:com1} and Table~\ref{tab:com2}, our proposed method has higher ASR than the baselines for all six datasets, especially when the poison ratio is small. For example, we can achieve ASR of 94.5\%, 44.69\%, 90.88\%, 96.53\% on VGG-Flower-l, CIFAR-10, GTSRB, CIFAR-100 models at poison ratios of 10\%, 1\%, 0.3\%, 0.1\% respectively, while BadNets only reaches ASR of 22.0\% (VGG-Flower-l), 10.00\% (CIFAR-10), 22.01\% (GTSRB), 1.29\% (CIFAR-100). Compared with HB that uses invisible triggers, we can achieve a significantly higher ASR across all datasets at all poison ratios.
{\color{black}For the high-resolution datasets, we can achieve an ASR of 88.82\% and 83.00\% on VGG-Flower-h and ImageNette at only 5\% and 15\% poison ratio, which is much higher than the baselines, especially BadNets, TrojanNN, and HB. }
Moreover, we can maintain a high CDA.

We compare the invisibility of the backdoored samples across all attacks, as shown in Fig.~\ref{fig:poisoned}. We can see that except for HB and ours, the triggers of all other baselines are conspicuous and easily detected by human eyes. Compared with HB, we can produce more indiscernible triggers in some cases. HB can not achieve a high ASR as ours. 

\subsection{{\color{black}Comparison with Baselines against ViT}}

{\color{black}We compared our proposed method with state-of-the-art vision transformer backdoor attacks, namely DBIA \cite{lv2021dbia}, DBAVT \cite{doan2023defending}, BAVT \cite{subramanya2022backdoor}, and TrojViT \cite{zheng2023trojvit}.
To implement the baseline attacks, we utilized their published source codes. 

The baselines and our attacks employed a default trigger size of $16\times 16$ and a default poisoning rate of 3\%. As demonstrated in Table~\ref{tab:Comvit}, our proposed attack method consistently outperforms the baselines across all six datasets, particularly in terms of the image quality metric LPIPS. The significantly lower LPIPS values achieved by our method (0.0122, 0.011, 0.0161, 0.0117, 0.0124, and 0.0101 for the six datasets, respectively) indicate that the backdoored samples generated by our method exhibit greater naturalness. Additionally, our method maintains a high prediction accuracy on clean samples.}

{\color{black}
We also present the backdoored samples of both our proposed method and the baselines across all attacks, as shown in Fig.~\ref{fig:poisoned-vit}. It is evident that, apart from BAVT and our proposed attack, the triggers in all other baselines are visible to the human eye and easily detectable. Since BAVT is based on the HB attack, which is also a hidden backdoor attack, its backdoored samples appear natural. However, we demonstrate that its attack performance is significantly lower than ours.}

\subsection{Ablation Study}
{\color{black}The ablation study results are shown in shown in Table \ref{tab:attention2} and Table \ref{tab:abtransformer}. 
The ``Base'' attack is a traditional backdoor attack with square-shaped model-dependent triggers placed at the bottom right corner of the image. The ``Base+Attn'' attack uses the attention mechanism to determine the trigger mask. The ``Base+Attn+Iter'' attack iteratively updates the trigger and the backdoored model using the co-optimization attack framework. The ``All'' attack is the complete attack with attention-based mask determination, co-optimization, and alternating retraining strategies. 

\textbf{Ablation study for DNN models.}
The ablation results for DNN models are shown in Table~\ref{tab:attention2}. Comparing ``Base" and ``Base+Attn", we can observe that the attention mechanism can significantly improve ASR, especially when the trigger is very small. 
As the trigger size becomes larger, the difference in ASR between the ``Base'' attack and the ``Base+Attn'' attack shrinks as the ``Base'' attack has more chance to select the pixels of high importance. }

{\color{black}Compared with the ``Base+Attn'' attack, the ``Base+Attn+Iter'' attack further increases ASR. We can observe that co-optimization improves both ASR and CDA. The alternating retraining strategy primarily improves the prediction accuracy of clean samples. Although experiments show that the attack success rate may slightly decrease at times, this reduction is negligible compared to the increase in the prediction accuracy of the backdoored model. For instance, ``Base+Attn+Iter' can yield a prediction accuracy of 89.07\% and an attack success rate of 99.62\% using the traditional retraining strategy in the CIFAR-10 dataset with a trigger size of $3\times 3$, while the alternating retraining strategy reaches 90.23\% prediction accuracy and 99.56\% attack success rate. However, in most cases, we discovered that the attack success rate would not decrease.

\textbf{Ablation study for ViT.}
As presented in Table~\ref{tab:abtransformer}, the ablation results for ViT exhibit a similar pattern of regularity as it does for CNN.

When comparing the performance of ``Base" and ``Base+Attn," it is evident that the attention mechanism brings about substantial enhancements in ASR, particularly for the CIFAR-10 and GTSRB datasets. Furthermore, the ``Base+Attn+Iter" attack achieves a higher ASR than the ``Base+Attn" attack. This observation highlights the positive impact of co-optimization on both ASR and CDA. In terms of the alternating retraining strategy, it primarily enhances the prediction accuracy of clean samples. However, in some cases, it results in a slight decrease in the attack success rate.}

\subsection{{\color{black}Impact of neuron gradient boosting}}
 In this part, we explore neuron gradient boosting on our attack performance against both DNN models and ViT models.
The results are shown in Table~\ref{tab:neuron-dnn} and Table~\ref{tab:neuron-vit}. 

We can see that the neuron gradient boosting strategy can significantly improve the attack success rate and clean data accuracy across all datasets and model types. For example, the gradient boosting strategy can achieve an ASR of 99.86\% and a CDA of 89.74\% for the CIFAR-10 dataset against the ViT model, while we can only achieve an ASR of 88.36\% and a CDA of 79.41\% without the neuron gradient boosting strategy. {\color{black}The possible reason is that the key neurons play an important part in classifying into the target label, henceforth enhancing its gradient causes the model to reach a better attack effect.}

\subsection{{\color{black}Impact of layer selection in ViT}}

{\color{black}The transformer model consists of patch embedding, attention layers, and the head. After analyzing these parts of the ViT in Section~\ref{sec:4}, we chose to select the neuron within the head layer. In this section, we evaluate the impact of layer selection on attack performance. For attention layers, we select the neuron from the class token, i.e., the first token, which is widely used as an explainable part of ViTs \cite{abnar2020quantifyingattentionflowtransformers, Chefer_2021_CVPR}. The results are shown in Table~\ref{tab:residing layer}. 

Compared to patch embedding and attention layers, the head layer demonstrates the best performance among all layers. Furthermore, posterior attention layers, such as layer 8 and layer 11, perform significantly better than prior layers. This aligns with the explainability in transformers \cite{abnar2020quantifyingattentionflowtransformers}, indicating that neurons from posterior layers are more representative and correlated with the target label. 

The superior performance of the head layer can be attributed to its structure. The head consists of a fully connected layer that links each input to each output through weighted connections. This configuration allows neurons in the head layer to respond strongly to the input trigger and the output results.

}

\section{Evading State-of-the-Art Backdoor Defenses}\label{sec:defense}

\subsection{Evading DNN-specific Backdoor Defenses}
{\color{black}We explore whether we can evade state-of-the-art backdoor defenses, including model pruning, NAD \cite{li2021neuralv}, STRIP \cite{gao2019strip}, and MNTD \cite{xu2019detecting}. For baseline attacks, we adjust the poison ratio as the default poison ratio is ineffective in certain cases. In particular, we set the poison ratio as 30\% in all baselines for VGG-Flower-l and VGG-Flower-h. We set the poison ratio as 20\% in HB for CIFAR-10. We set the poison ratio as 3\% in BadNets, TrojanNN, and HB for CIFAR-100, and 30\% in BadNets, TrojanNN, and HB for ImageNette. Others adopt the default poison ratio.}

\subsubsection{Model Pruning}
The defender first ranks neurons in ascending order according to the average activation by clean samples. Then, the defender sequentially prunes neurons until the accuracy of the validation dataset drops below a predetermined threshold.

As shown in Fig.~\ref{fig:prune}, we can still achieve high ASR after pruning. Given a threshold of 80\% for CDA, we can preserve an ASR of more than 82\% for all datasets. This means that we are resistant to model pruning.


\subsubsection{NAD} 
In NAD \cite{li2021neuralv}, the defender first fine-tunes the backdoored model on a small set of benign samples and uses the fine-tuned model as a teacher model. Then, NAD uses the teacher model to distill the backdoored model (student model) through attention distillation. In this way, the neurons of the backdoor will be aligned with benign neurons associated with meaningful representations.

As shown in Table~\ref{tab:nad}, after applying NAD, the ASR of ours only slightly decreases. The possible reason is that the gap between our generated backdoored model and the benign model has been narrowed through alternating retraining.


\subsubsection{STRIP} 
In STRIP, the defender duplicates an input sample for many times and merges each copy with a different sample to generate a set of perturbed samples. The distribution of the prediction results of the perturb samples is used to detect backdoored samples.
It is assumed that the prediction results of the disturbed samples have a high entropy if the sample is clean and a low entropy if the sample contains the trigger as the trigger strongly drives the prediction results toward the target label.

As shown in Fig.~\ref{fig:STRIP}, the prediction results of our backdoored samples have a similar entropy distribution to benign samples for all datasets, making it difficult to differentiate the backdoored samples and the benign samples. Thus, we can evade STRIP defense.

\subsubsection{MNTD} \label{sec:MNTD}

MNTD \cite{xu2019detecting} is a model-based defense based on a binary meta-classifier. To train the meta-model, the defender builds a large number of benign and backdoored shadow models as training samples. Since the defender has no knowledge of the specific backdoor attack methods, MNTD adopts \emph{jumbo learning} to generate a variety of backdoored models. In this way, MNTD is generic and can detect most state-of-the-art backdoor attacks. 
To apply MNTD to our attack framework, for each dataset, we generate 2,048 benign models and 2,048 backdoored models to train a well-performed meta-classifier. 

When we feed our backdoored models to the meta-classifier, it is shown that they can all evade the inspection of MNTD. In comparison, when we feed the backdoored models of the baselines to the meta-classifier, they are all detected by MNTD.
The success in evading the detection of MNTD is possibly due to our alternating retraining strategy that makes the backdoored models behave like the benign ones. 

\subsubsection{NC} 
{\color{black}
NeuralCleanse (NC) \cite{wang2019neural} employs a model-based defense strategy that aims to recover triggers by calculating the minimal perturbation required for a sample with the source label to be misclassified as the target label. The target label requiring the smallest perturbation is identified as the actual target, with this perturbation considered the trigger.

The recovered triggers and the corresponding real triggers are shown in Fig. \ref{fig:NC}. We can see a significant discrepancy between our generated triggers and the recovered ones. Additionally, we observe that NC's reversed triggers on high-resolution data samples are more dispersed and harder to identify. We then utilize Median Absolute Deviation (MAD) for anomaly detection, with a threshold set at 2. Experimental results consistently show that the MAD values for our target classes remain below this threshold (0.5415 for VGG-Flower-l, 0.0920 for CIFAR-10, 0.5040 for GTSRB, 1.0672 for CIFAR-100, 0.8313 for ImageNette, and 1.7584 for VGG-Flower-h). The effectiveness of our proposed attack may be attributed to its opacity adjustment, which makes it more challenging to recover low-magnitude triggers.
}

\subsubsection{ABS} 
{\color{black}
ABS \cite{liu2019abs} is a model-based defense method designed to detect backdoored models by analyzing neuron behaviors. It identifies potentially compromised neurons by stimulating them and observing changes in output, followed by optimization to reverse-engineer the backdoor triggers. ABS achieves a detection rate exceeding 90\% even with a limited number of input samples, proving effective across various datasets and model architectures.

In the experiments, we deploy ABS during the neuron selection stage to detect and deactivate compromised neurons. The experimental results demonstrate that our proposed attack can successfully bypass ABS's defense mechanisms. Despite the application of ABS, we achieve attack success rates of 94.93\% for VGG-Flower-l, 97.67\% for CIFAR-10, 99.03\% for GTSRB, 97.10\% for CIFAR-100, 95.00\% for ImageNette, and 93.88\% for VGG-Flower-h. This success can be attributed to the natural and stealthy design of our triggers. By adjusting transparency and employing QoE-based triggers, we alter the distribution of neuron activations during model training, rather than merely activating a few neurons abnormally.
}

\subsection{{\color{black}Evading ViT-Specific Backdoor Defenses}}

{\color{black}Currently, there are only a limited number of available defenses specifically designed for Vision Transformers (ViTs). In this study, we assess the robustness of our proposed attack method against DBAVT \cite{doan2023defending}, which represents the most advanced ViT-Specific backdoor defense. DBAVT mitigates backdoor attacks on ViTs by employing patch processing. It is based on the insight that the accuracy of clean data and the success rates of backdoor attacks in ViTs respond differently to patch processing before positional encoding, unlike in CNN models.

We applied DBAVT to both our proposed attack and baseline attacks, and the results are shown in Table~\ref{tab:dbavt}. It is demonstrated that even after applying DBAVT, we maintain a high attack success rate (ASR). For VGG-Flower-l, CIFAR-10, GTSRB, CIFAR-100, ImageNette, and VGG-Flower-h, we achieve ASRs of 73.5\%, 83.47\%, 90.67\%, 89.73\%, 88.45\%, and 72\%, respectively. The possible reason is that to maintain a high prediction accuracy of the model, the percentage of patches dropped and shuffled of DBAVT is limited when defending against
our method. Therefore, our proposed attack framework demonstrates robustness against DBAVT.

In terms of the baselines, BAVT also shows resilience to the defense, maintaining a high ASR. In contrast, other baseline attacks see their ASRs reduced to less than 60\% in most cases. Specifically, the DBAVT attack is especially susceptible to this defense, reducing the ASR to less than 41\%. Note that in \cite{doan2023defending}, the authors proposed both an attack and a defense.
}

\section{Conclusion}
This paper presents the design, implementation, and evaluation of an effective and evasive backdoor attack against deep neural networks and vision transformers. To obtain the effectiveness goal, we proposed a novel attention-based mask generation strategy and utilized a co-optimized attack framework. To achieve the evasiveness goal, we carefully adjust the trigger transparency and add a QoE constant to the loss function. We also propose an alternating retraining strategy to improve the model prediction accuracy. We show that our proposed attacks can evade state-of-the-art backdoor defenses.
Experiments on VGG-Flower, GTSRB, CIFAR-10, CIFAR-100, and ImageNette verify the superiority of the attack when compared with state-of-the-art backdoor attacks. 

\bibliographystyle{plain}
\bibliography{main}

\begin{IEEEbiography}[{\includegraphics[width=1in,height=1.25in,clip,keepaspectratio]{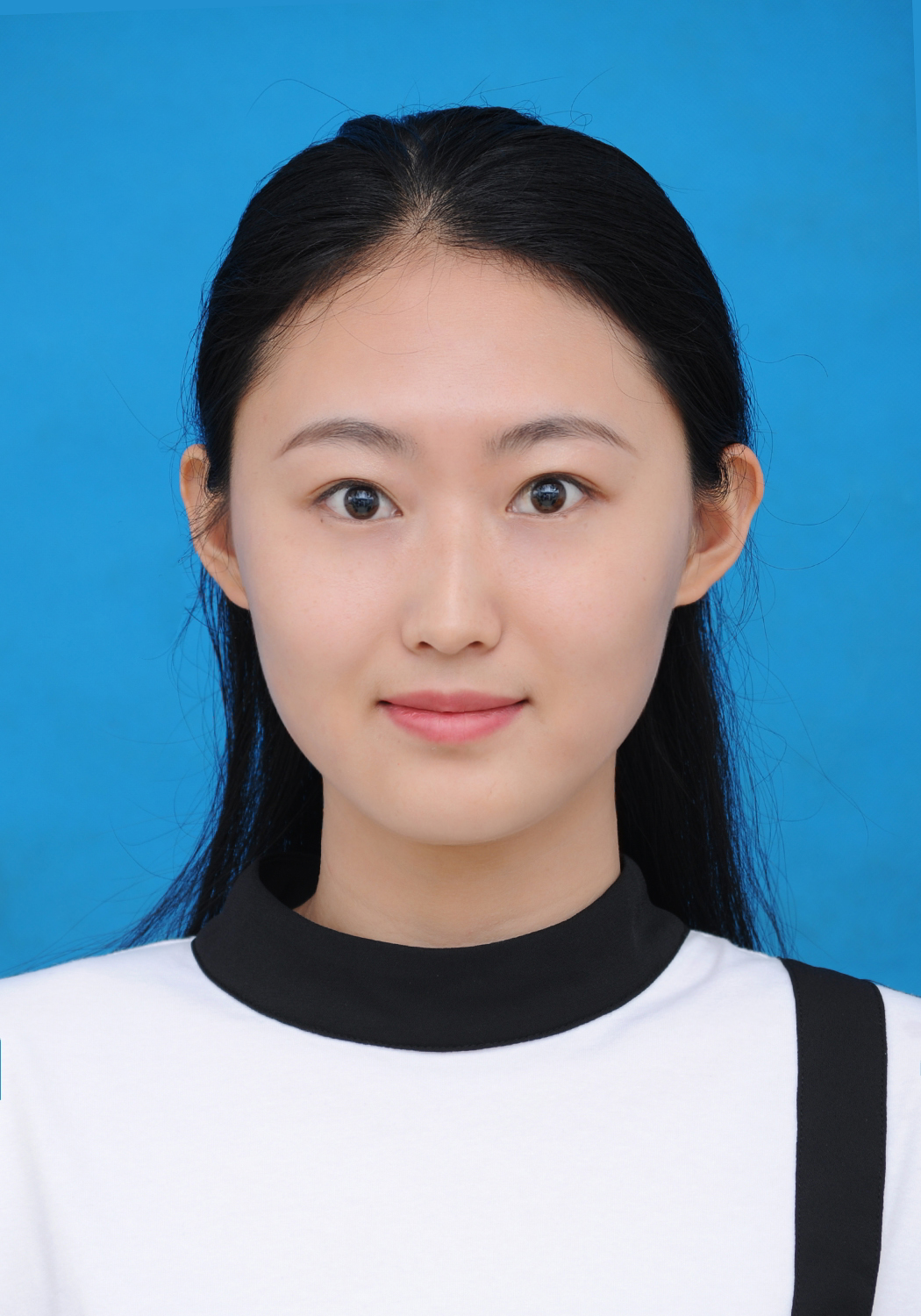}}]{Xueluan Gong} received her B.S. degree in Computer Science and Electronic Engineering from Hunan University in 2018. She received her Ph.D. degree in Computer Science from Wuhan University in 2023. She is currently a Research Fellow at the School of Computer Science and Engineering at the Nanyang Technological University, Singapore.
Her research interests include network security, AI security, and data mining. She has published more than 30 publications in top-tier international journals or conferences, including IEEE S\&P, NDSS, ACM CCS, Usenix Security, WWW, ACM Ubicomp, IJCAI, IEEE JSAC, TDSC, TIFS, etc.
\end{IEEEbiography}

\begin{IEEEbiography}[{\includegraphics[width=1in,height=1.2in,clip,keepaspectratio]{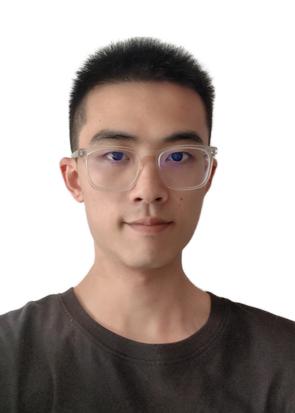}}]{Bowei Tian} is currently pursuing the B.E. at the School of Cyber Science and Engineering from Wuhan University, China. His research interests include network security and information security.
\end{IEEEbiography}

\begin{IEEEbiography}[{\includegraphics[width=1in,height=1.25in,clip,keepaspectratio]{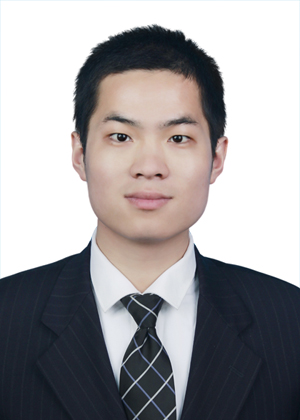}}]{Meng Xue} received his Ph.D. degree in Computer Science School from Wuhan University in 2022. He is currently a Postdoc in the Department of Computer Science and Engineering at Hong Kong University of Science and Technology. His research interests include the Internet of Things, smart sensing, and AI security.
\end{IEEEbiography}

\begin{IEEEbiography}[{\includegraphics[width=1in,height=1.25in,clip,keepaspectratio]{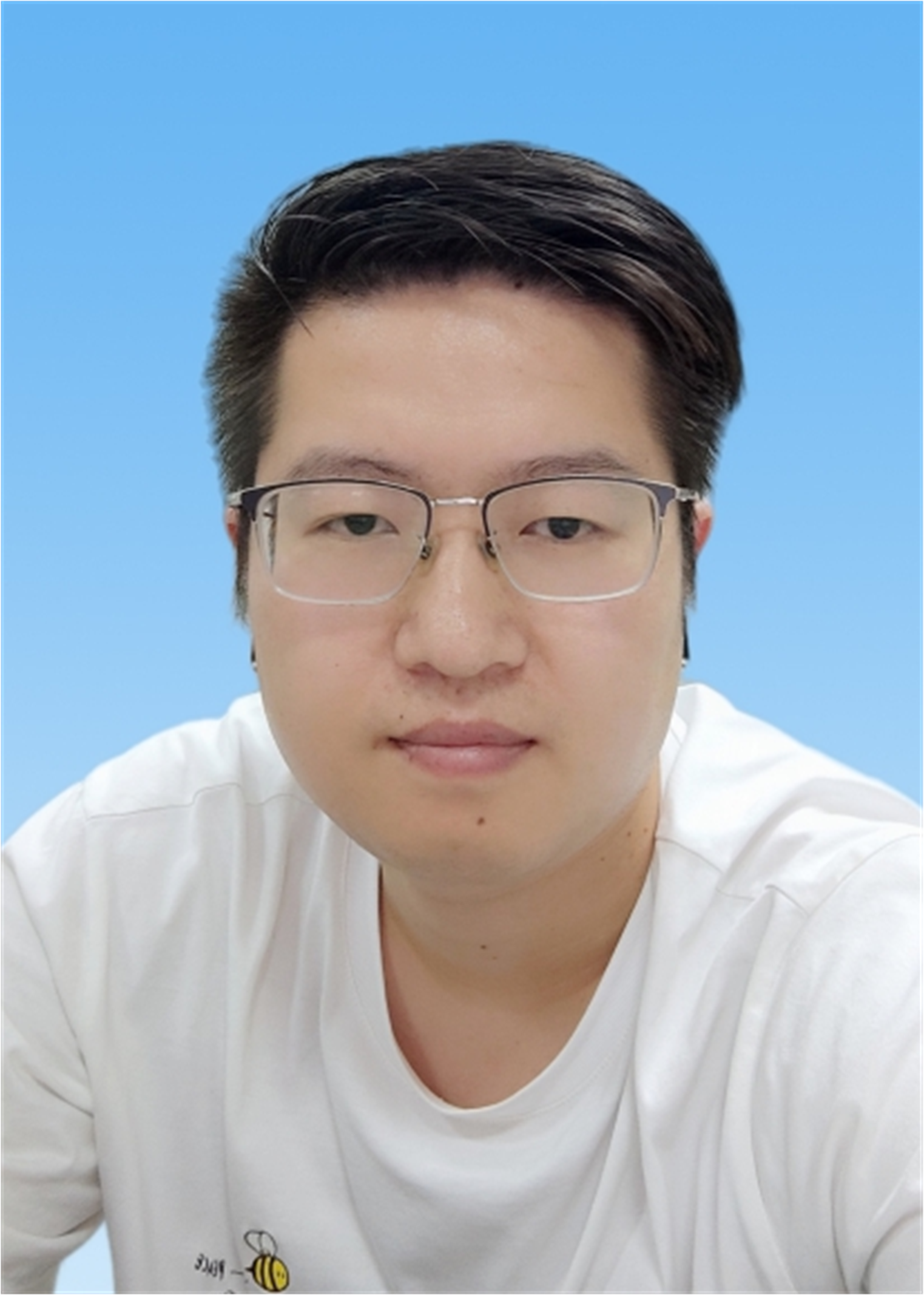}}]{Yuan Wu} received his Ph.D. degree in the School of Computer Science, Wuhan University. He is currently an Associate Professor in the School of Computer Science and Artificial Intelligence at Wuhan Textile University. His research interests include mobile sensing, the Internet of Things, and wearable devices.
\end{IEEEbiography}

\begin{IEEEbiography}[{\includegraphics[width=1in,height=1.25in,clip,keepaspectratio]{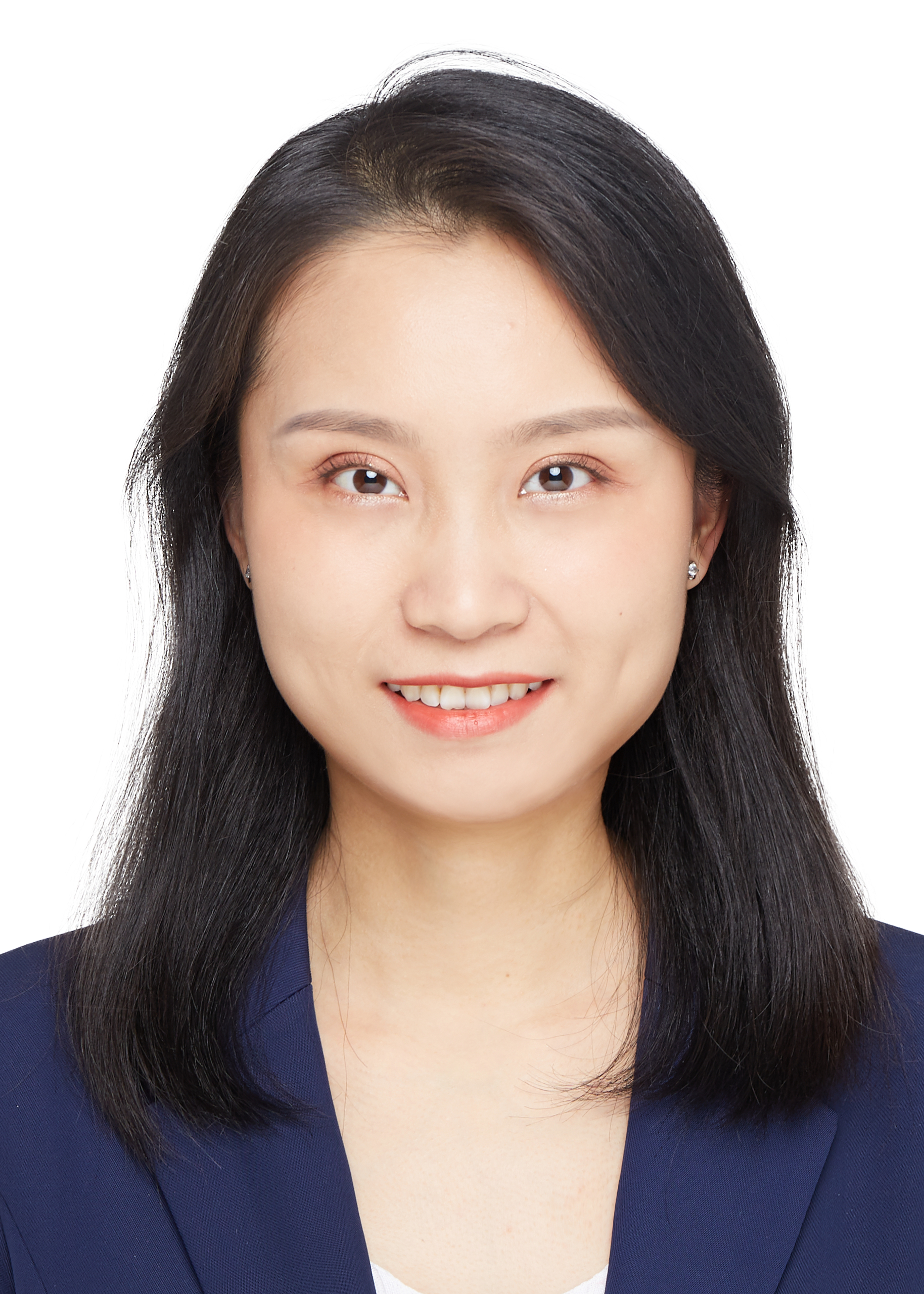}}]{Yanjiao Chen} received her B.E. degree in Electronic Engineering from Tsinghua University in 2010 and Ph.D. degree in Computer Science and Engineering from Hong Kong University of Science and Technology in 2015. She is currently a Bairen researcher in Zhejiang University, China. Her research interests include spectrum management for Femtocell networks, network economics, network security, AI security, and Quality of Experience (QoE) of multimedia delivery/distribution.
\end{IEEEbiography}

\begin{IEEEbiography}[{\includegraphics[width=1in,height=1.25in, clip,keepaspectratio]{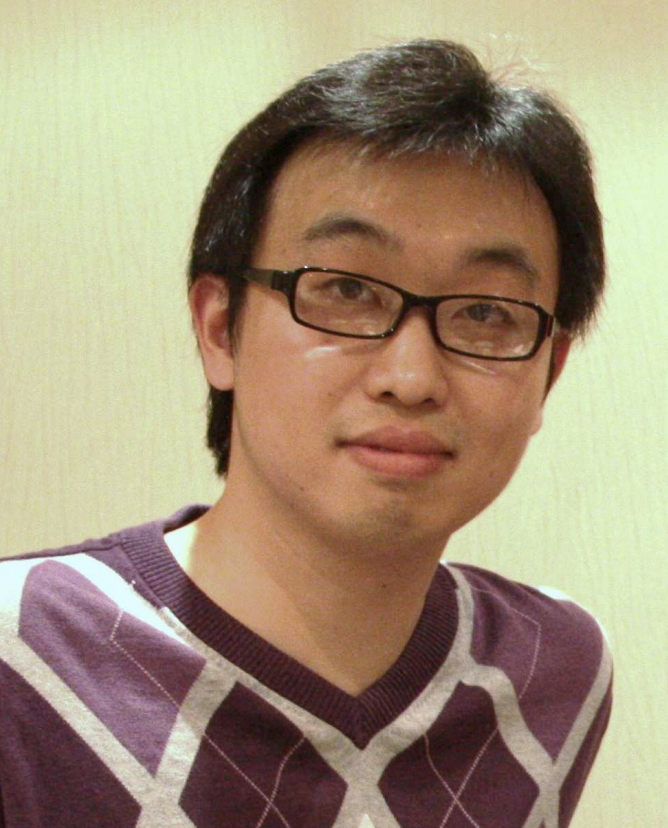}}]{Qian Wang} is a Professor in the School of Cyber Science and Engineering at Wuhan University, China. He was selected into the National High-level Young Talents Program of China, and listed among the World's Top 2\% Scientists by Stanford University. He also received the National Science Fund for Excellent Young Scholars of China in 2018. He has long been engaged in the research of cyberspace security, with focus on AI security, data outsourcing security and privacy, wireless systems security, and applied cryptography. He was a recipient of the 2018 IEEE TCSC Award for Excellence in Scalable Computing (early career researcher) and the 2016 IEEE ComSoc Asia-Pacific Outstanding Young Researcher Award. He has published 200+ papers, with 120+ publications in top-tier international conferences, including USENIX NSDI, ACM CCS, USENIX Security, NDSS, ACM MobiCom, ICML, etc., with 20000+ Google Scholar citations. He is also a co-recipient of 8 Best Paper and Best Student Paper Awards from prestigious conferences, including ICDCS, IEEE ICNP, etc. In 2021, his PhD student was selected under Huawei's  ``Top Minds'' Recruitment Program. He serves as Associate Editors for IEEE Transactions on Dependable and Secure Computing (TDSC) and IEEE Transactions on Information Forensics and Security (TIFS). He is a fellow of the IEEE, and a member of the ACM.
\end{IEEEbiography}